\documentclass[sigconf, 9pt, screen]{acmart}

\usepackage{subcaption}
\usepackage{todonotes}
\usepackage{url}
\usepackage{enumitem}
\usepackage{graphicx}
\usepackage{caption}
\usepackage{amsmath}
\usepackage{tabu}
\usepackage{multirow}
\usepackage{hyperref}
\usepackage{adjustbox}
\usepackage{amsfonts}
\usepackage{algpseudocode}
\usepackage{pgfplots}
\pgfplotsset{compat=1.12}
\usepackage{soul}
\usepackage{footnote}
\usepackage{lipsum}
\usepackage[ruled,vlined, linesnumbered]{algorithm2e}
\usepackage{subcaption}
\usepackage{listings}
\usepackage{tikz}
\colorlet{punct}{red!60!black}
\definecolor{background}{HTML}{FFFFFF}
\definecolor{delim}{RGB}{20,105,176}
\colorlet{numb}{magenta!60!black}

\colorlet{punct}{red!60!black}
\definecolor{background}{HTML}{FFFFFF}
\definecolor{delim}{RGB}{20,105,176}
\colorlet{numb}{magenta!60!black}

\newcommand{\shrinkspace}{\vspace{-5mm}}
\newcommand*\circled[1]{\tikz[baseline=(char.base)]{
            \node[shape=circle,draw,inner sep=0.75pt, text=white,fill=black] (char) {#1};}}
\newcommand*\circledw[1]{\tikz[baseline=(char.base)]{
            \node[shape=circle,draw,inner sep=0.75pt, text=black,fill=white] (char) {#1};}}
\lstdefinelanguage{json}{
    basicstyle=\footnotesize\ttfamily,
    numbers=left,
    numberstyle=\scriptsize,
    xleftmargin=2.3em,
    xrightmargin=0.5em,
    framexleftmargin=1.9em,
    stepnumber=1,
    numbersep=8pt,
    showstringspaces=false,
    breaklines=true,
    frame=single,
    backgroundcolor=\color{background},
    literate=
     *{0}{{{\color{numb}0}}}{1}
      {1}{{{\color{numb}1}}}{1}
      {2}{{{\color{numb}2}}}{1}
      {3}{{{\color{numb}3}}}{1}
      {4}{{{\color{numb}4}}}{1}
      {5}{{{\color{numb}5}}}{1}
      {6}{{{\color{numb}6}}}{1}
      {7}{{{\color{numb}7}}}{1}
      {8}{{{\color{numb}8}}}{1}
      {9}{{{\color{numb}9}}}{1}
      {:}{{{\color{punct}{:}}}}{1}
      {,}{{{\color{punct}{,}}}}{1}
      {\{}{{{\color{delim}{\{}}}}{1}
      {\}}{{{\color{delim}{\}}}}}{1}
      {[}{{{\color{delim}{[}}}}{1}
      {]}{{{\color{delim}{]}}}}{1},
}

\copyrightyear{2024}
\acmYear{2024}
\setcopyright{acmlicensed}\acmConference[SAC '24]{The 39th ACM/SIGAPP Symposium on Applied Computing}{April 8--12, 2024}{Avila, Spain}
\acmBooktitle{The 39th ACM/SIGAPP Symposium on Applied Computing (SAC '24), April 8--12, 2024, Avila, Spain}
\acmPrice{15.00}
\acmDOI{10.1145/3605098.3636015}
\acmISBN{979-8-4007-0243-3/24/04}

\begin{document}

\title{Training Heterogeneous Client Models using Knowledge Distillation in Serverless Federated Learning}



\author{Mohak Chadha$^{1}$, Pulkit Khera$^{1}$, Jianfeng Gu$^{1}$, Osama Abboud$^{2}$, Michael Gerndt$^{1}$ 
}
\affiliation{%
\institution{$^1$\{firstname.lastname\}@tum.de, Technische Universit{\"a}t M{\"u}nchen 
\country{Germany}}
}
\affiliation{%
\institution{$^2$\{firstname.lastname\}@huawei.com, Huawei Technologies
\country{Germany}}
}

\renewcommand{\shortauthors}{Mohak Chadha et al.}

\begin{abstract}
Federated Learning (FL) is an emerging machine learning paradigm that enables the collaborative training of a shared global model across distributed clients while keeping the data decentralized. Recent works on designing systems for efficient FL have shown that utilizing serverless computing technologies, particularly Function-as-a-Service (FaaS) for FL, can enhance resource efficiency, reduce training costs, and alleviate the complex infrastructure management burden on data holders. However, existing serverless FL systems implicitly assume a uniform global model architecture across all participating clients during training. This assumption fails to address fundamental challenges in practical FL due to the resource and statistical data heterogeneity among FL clients. To address these challenges and enable heterogeneous client models in serverless FL, we utilize Knowledge Distillation (KD) in this paper. Towards this, we propose novel optimized serverless workflows for two popular conventional federated KD techniques, i.e., \texttt{FedMD} and \texttt{FedDF}. We implement these workflows by introducing several extensions to an open-source serverless FL system called \emph{FedLess}. Moreover, we comprehensively evaluate the two strategies on multiple datasets across varying levels of client data heterogeneity using heterogeneous client models with respect to accuracy, fine-grained training times, and costs. Results from our experiments demonstrate that serverless \texttt{FedDF} is more robust to extreme non-IID data distributions, is faster, and leads to lower costs than serverless \texttt{FedMD}. In addition, compared to the original implementation, our optimizations for particular steps in \texttt{FedMD} and \texttt{FedDF} lead to an average speedup of $3.5$x and $1.76$x across all datasets.



\end{abstract}

%
%
\vspace{-2mm}

\vspace{-2mm}


\maketitle
\vspace{-4mm}
\section{Introduction}
\label{sec:intro}
Increasing concerns about data privacy and recent legislations such as the Consumer Privacy Bill of Rights in the U.S.~\cite{Privacy_and_big_data} prevent the training of ML models using the traditional centralized learning approach~\cite{lecun2015deep}.  With the goal of not exposing raw data as in centralized learning, an emerging distributed training paradigm called Federated Learning (FL)~\cite{mcmahan2017communication} has gained significant popularity in various application domains, such as banking~\cite{ludwig2020ibm} and mobile services~\cite{huba2022papaya}. 

FL enables the collaborative training of a shared global ML model across remote devices or \texttt{clients} while keeping the training data decentralized. The traditional FL training process~\cite{mcmahan2017communication} is \textit{synchronous} and occurs in multiple rounds. A main component called the \texttt{central} \texttt{server} organizes the training process and decides which clients contribute in a new round. During each round, clients improve the shared global model by optimizing it on their local datasets and sending back only the updated model parameters to the central server. Following this, the local model updates from all participating clients are collected and aggregated to form the updated consensus model. Recent works on designing systems for efficient FL have shown that both components in an FL system, i.e., the \texttt{clients} and the \texttt{central} \texttt{server}, can immensely benefit from an emerging cloud computing paradigm called \textit{serverless computing}~\cite{serverlessfl, fedless, fedlesscan, jayaram2022lambda, jitfl, jayaramadaptive, flox}.

Function-as-a-Service (FaaS) is the computational concept of serverless computing and has gained significant popularity and widespread adoption in various application domains such as machine learning~\cite{fedless, fedlesscan, serverlessfl, fastgshare}, edge computing~\cite{fado}, heterogeneous computing~\cite{fncapacitor, jindal2021function, courier, postericdcs}, and scientific computing~\cite{chadha2021architecture, demystifying}. In FaaS,  developers implement fine-grained pieces of code called \textit{functions} that are packaged independently in containers and uploaded to a FaaS platform. These functions are \textit{ephemeral}, i.e., short-lived, and \textit{event-driven}, i.e., these functions only get executed in response to external triggers such as HTTP requests. Moreover, these functions are \textit{stateless}, i.e., any application state needs to be persisted in external storage. Several open-source and commercial FaaS platforms, such as OpenFaaS~\cite{openfaas} and Google Cloud Functions (GCF)~\cite{gcloud-functions-2}, are currently available. In serverless FL, clients are independent functions deployed onto a FaaS platform and capable of performing their model updates. 

\begin{figure}[t]
\centering
\includegraphics[width=0.70\columnwidth]{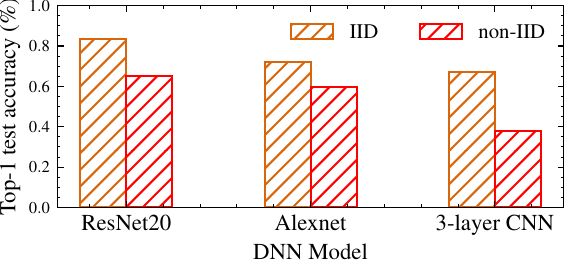}
\vspace{-2mm}
\caption{Global model accuracy across different model architectures on the CIFAR-10 dataset with \texttt{FedAvg}~\cite{mcmahan2017communication} for both IID and non-IID data distributions. The models are trained with \emph{FedLess}~\cite{fedless} for $100$ clients.}
\label{fig:motivation_example}
\shrinkspace
\vspace{-3mm}
\end{figure}

Most standard stateful FL systems~\cite{flower, oort} and all current serverless FL systems implicitly assume that all FL clients must have a uniform ML model architecture to train a global consensus model. However, a practical FL system is affected by different fundamental client-level challenges that are difficult to address with this assumption. These include computational heterogeneity and statistical data heterogeneity. FL clients in the wild~\cite{huba2022papaya} can vary from small edge devices to high-performant GPU-enabled systems with varying memory, compute, and storage capacities. Therefore, it is not always feasible for each participating client to agree on a single global model architecture. Opting for a simple architecture may restrict the capacity of the collaboratively trained global model, whereas selecting a large and complex model architecture can substantially increase the duration required for FL training~\cite{fedlesscan}. In addition, clients in practical FL systems have \textit{unbalanced non-IID} data distributions, i.e., the private data samples held by individual clients exhibit variations in their statistical properties, such as feature distributions, class imbalances, or data biases~\cite{hsieh2020non}. As a result, in extreme non-IID scenarios, the uniform global model may lead to poor generalization performance following the model aggregation process due to high variance among the trained client models. Figure~\ref{fig:motivation_example} shows the impact of non-IID data distribution among FL clients on the accuracy of the trained global model on the CIFAR-10 dataset. Across all model architectures, we observe a $27.1$\% average decrease in the accuracy of the trained global model with \texttt{FedAvg}~\cite{mcmahan2017communication}.  To this end, each FL client needs to have the flexibility to choose their own model architecture personalized to their private data distribution while simultaneously benefiting from other collaborating clients.

To address these challenges in practical FL systems,  we utilize Knowledge Distillation (KD)~\cite{hinton} in this paper. KD is a popular technique used in ML that facilitates the transfer of knowledge from a large and complex model, known as the \textit{teacher model}, to a smaller and more efficient model, referred to as the \textit{student model}. KD enables heterogeneous client models in FL since knowledge transfer is achieved by distilling model prediction probabilities or  \texttt{logits} instead of directly exchanging model parameters between the student and teacher models. 
In addition, KD supports a high level of flexibility in the choice of client model architectures in contrast to other approaches like parameter decoupling~\cite{param_decoupling} that tolerate flexibility in only particular layers of the overall model architecture.

Most KD strategies in FL, such as \texttt{FedMD}~\cite{fedmd} and \texttt{MHAT}~\cite{mhat}, typically involve a series of steps, such as private training, prediction on public datasets, and aligning the local client models to the obtained logits. However, the synchronous nature of these algorithms introduces inefficiencies where the central server must wait for all the participating clients to complete a particular step before moving on to the subsequent steps. Due to the statistical and resource heterogeneity in FL, most clients in conventional KD-based FL systems remain idle, leading to resource wastage and unnecessary costs (Figure~\ref{fig:motivation_example_serverless}). Moreover, managing complex infrastructure for clients can be overwhelming for all data administrators. The FaaS computing model offers several advantages, such as no infrastructure management, automatic scaling to zero when resources are unused, and an attractive fine-grained \textit{pay-per-use} billing policy~\cite{rise}. Adapting and optimizing existing conventional KD-based FL techniques to utilize stateless FaaS technologies in both entities of an FL system can improve resource efficiency, reduce training costs, and address practical challenges in FL systems (Figure~\ref{fig:motivation_example_serverless}). To this end, our key contributions are:

\begin{figure}[t]
\centering
\includegraphics[width=0.75\columnwidth]{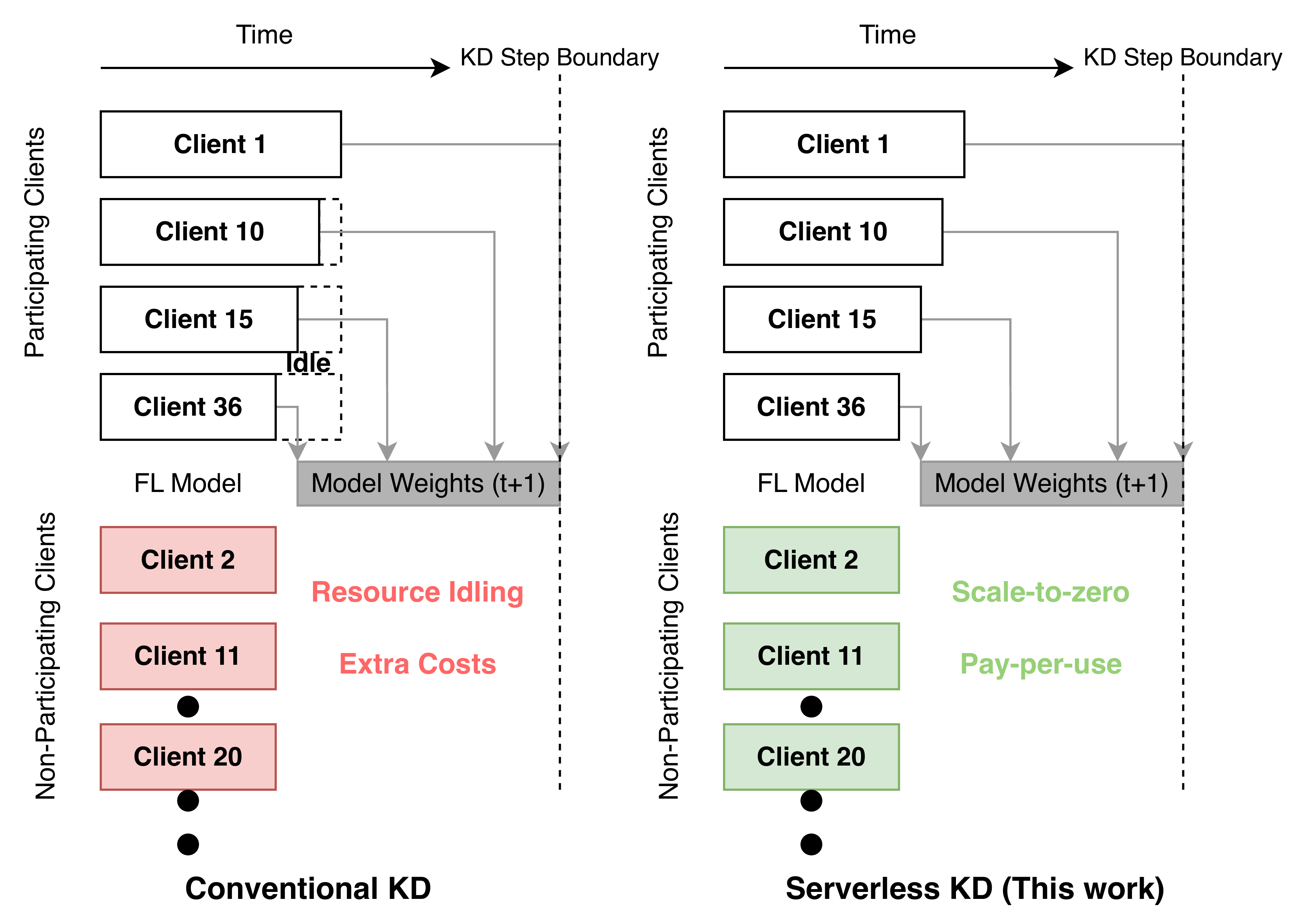}
\vspace{-2mm}
\caption{Synchronous conventional KD and serverless KD.}
\label{fig:motivation_example_serverless}
\shrinkspace
\end{figure}

\begin{itemize}
    \item We propose novel optimized serverless workflows for two popular conventional KD-based FL strategies, i.e., \texttt{FedMD} and \texttt{FedDF}. 
    \item We implement the two workflows by introducing several extensions to an open-source serverless FL system called \emph{FedLess}~\cite{fedless}\footnote{Our implementation can be found here: \url{https://github.com/Serverless-Federated-Learning/FedLess/tree/serverless-knowledge-distillation}.}. To the best of our knowledge, this represents the first system in the literature that supports training heterogeneous client models using serverless KD.
    \item We comprehensively evaluate the implemented strategies on multiple datasets across varying levels of client data heterogeneity using heterogeneous client model architectures wrt accuracy, fine-grained training times, and costs.
\end{itemize}

The rest of the paper is structured as follows. \S\ref{sec:background} provides an overview of KD in FL. In \S\ref{sec:relatedwork}, we describe the previous approaches related to our work. \S\ref{sec:sysdesign} describes our extensions to \emph{FedLess}. In \S\ref{sec:serverlesskd}, we describe the workflow of the serverless implementations of the two algorithms. \S\ref{sec:expsetup} describes our experimental setup, while in \S\ref{sec:expresults} our experimental results are presented. Finally, \S\ref{sec:concfuture} concludes the paper and presents an outlook.

\section{Background}
\label{sec:background}

\subsection{Knowledge Distillation in FL}
\label{sec:kdinfl}
When applied to FL, KD offers several advantages. These include model compression, heterogeneous and personalized client models, reduced communication overhead, and increased client privacy~\cite{kd_practical_guide}. Unlike traditional FL approaches~\cite{mcmahan2017communication} that require exchanging the entire model parameters, most KD-based techniques typically only involve the transmission of class scores. This reduces the communication overhead and minimizes the data transferred between clients and the server during model aggregation. In addition, this prevents \textit{model inferencing attacks}~\cite{enthoven2021overview} in FL. Towards this, several strategies that leverage KD in FL have been proposed. These strategies can be classified into four categories: \circledw{1} distillation of knowledge to each FL \texttt{client} to learn stronger personalized models, \circledw{2} distillation of knowledge to the \texttt{central server} to learn stronger server models, \circledw{3}  bidirectional distillation to both the FL \texttt{clients} and the \texttt{server}, and \circledw{4} \texttt{inter-client} distillation. 

\texttt{FedMD}~\cite{fedmd} belongs to the first category of KD algorithms, focusing on strengthening personalized models for each FL client. It requires a carefully selected labeled public dataset and offers flexibility in various learning tasks, including image and text data applications. In the second category of KD strategies, we considered two specific approaches, i.e.,  \texttt{MHAT}~\cite{mhat} and \texttt{FedDF}~\cite{feddf}. Both strategies focus on learning stronger (student) server models with the help of several (\emph{ensemble}) candidate (teacher) client models. For distilling knowledge to the server models, \texttt{MHAT} requires a labeled public dataset, while \texttt{FedDF} does not impose this restriction. \texttt{FedET}~\cite{fedet} is a bidirectional distillation technique that uses a weighted consensus algorithm with diversity regularization to train small client models and a large server model. Distributed distillation (\texttt{DD})~\cite{d_distillation} is a semi-supervised FL strategy where knowledge is transferred amongst all neighboring clients in a network. In each communication round,  clients exchange soft targets with all their neighbors and update their own targets using a consensus algorithm. Following this, the updated soft targets are used by each client to update its private model weights. 

Strategies belonging to categories three and four either require frequent communication between the server and the clients or amongst clients leading to significantly high communication costs. Therefore, we don't consider any distillation strategies from these categories for our serverless implementation. In this paper, we adapt and optimize the strategies \texttt{FedMD} and \texttt{FedDF} using FaaS functions. We choose \texttt{FedDF} over \texttt{MHAT} due to its superior robustness in handling heterogeneous data distributions and its flexibility regarding the type of public dataset required~\cite{feddf}.



\begin{figure}[t]
    \centering
    \includegraphics[width=0.85\columnwidth]{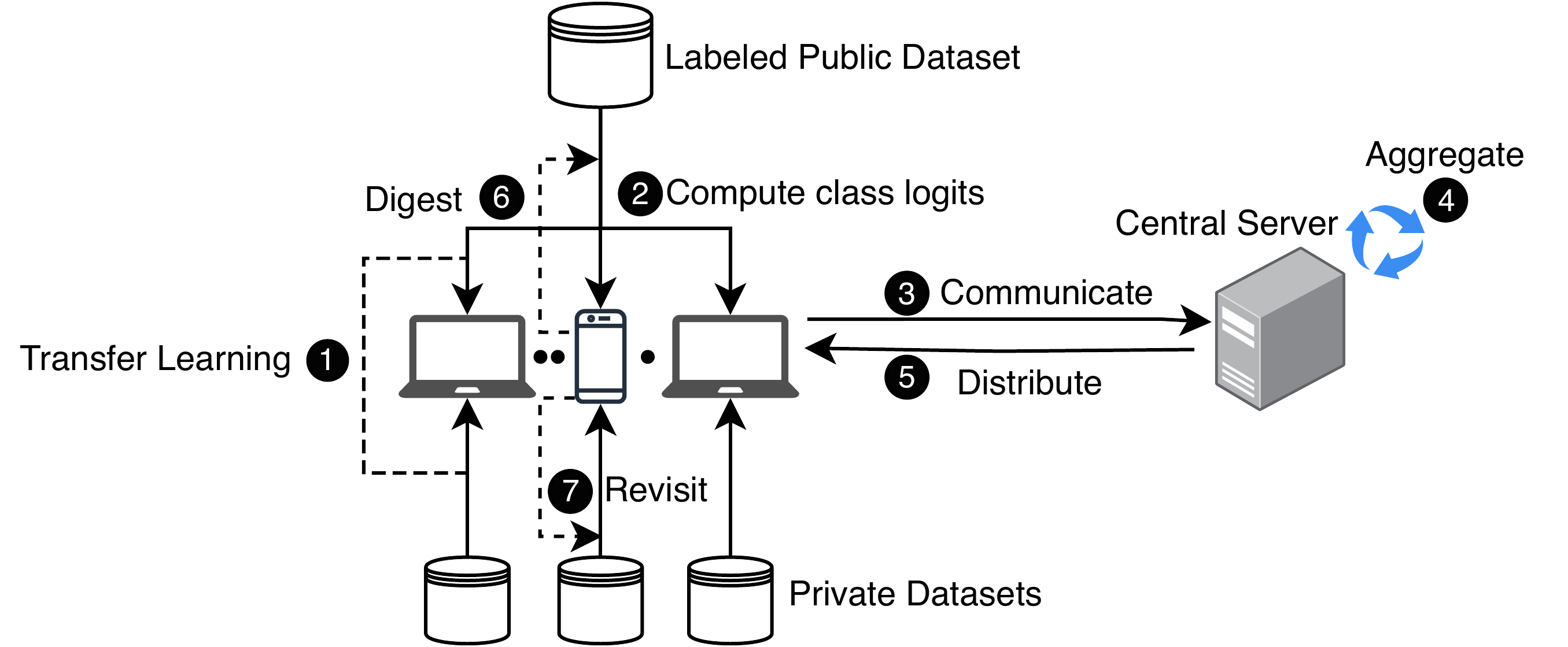}
    \vspace{-2mm}
    \caption{Different steps in the \texttt{FedMD}~\cite{fedmd} algorithm.} 
    \label{fig:fedmdworkflow}
    \shrinkspace
\end{figure}
\vspace{-2mm}
\subsection{FedMD}
\label{sec:femd}
The \texttt{FedMD}~\cite{fedmd} algorithm supports a unique model architecture for each participating client. In this strategy, the central server does not require any information about the client models and treats them as a black box. Furthermore, in addition to their private datasets, each client has access to a labeled public dataset. The collaborative training process in \texttt{FedMD} 
consists of seven synchronous steps, each of which is shown in Figure~\ref{fig:fedmdworkflow}. In the first step (\circled{1}), each participating client trains on the labeled public dataset until convergence and then on its private dataset for a few epochs. Following this, each participating client performs a forward-pass inference on the public dataset and sends the generated class logits to the central server (\circled{2}-\circled{3}). The central server aggregates the obtained class logits (\circled{4}) and sends them back to the participating clients (\circled{5}). After this, each participating client trains its model using the aggregated logits as soft targets on the public dataset (\circled{6}). Finally, each participating client trains the model on its private dataset for a few epochs for personalization (\circled{7}). While step \circled{1} only occurs during the start of the training process, steps \circled{2}-\circled{7} are repeated until the required client model accuracy is reached.  The final output layer of the client model architecture must have a total number of output neurons equal to the sum of the number of classes in the private dataset and the number of classes in the public dataset. This is because the model is trained on both datasets during the complete training process, and in most cases, the classes in the two datasets are mutually exclusive. Therefore, a good public dataset (\S\ref{sec:kdinfl}) minimizes the total number of output classes while maximizing the amount of available data.  
\vspace{-3mm}

\subsection{FedDF}
\label{sec:fedf}
\begin{figure}[t]
    \centering
    \includegraphics[width=\columnwidth]{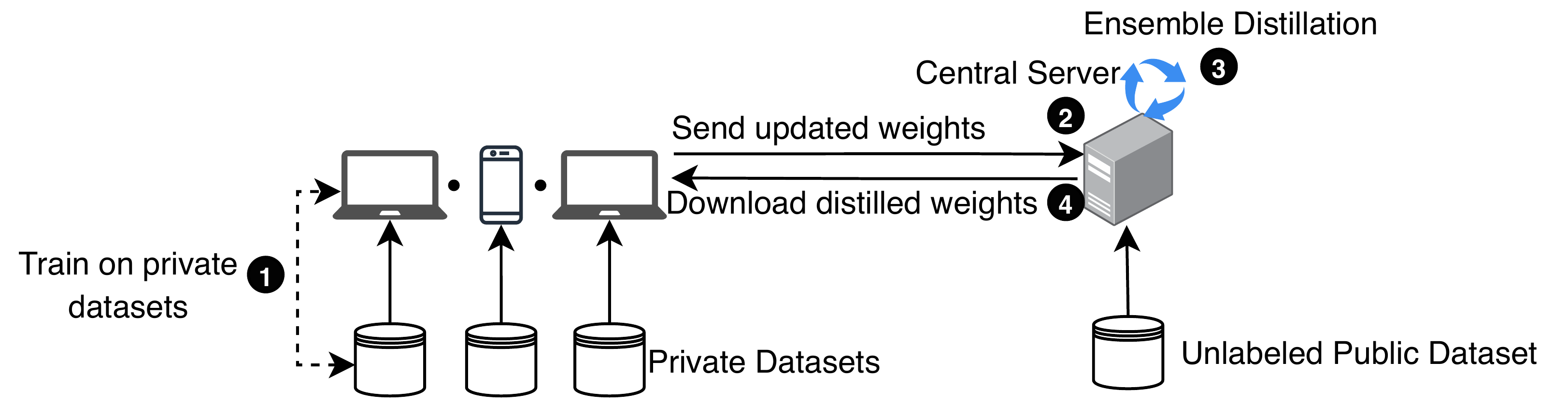}
    \vspace{-2mm}
    \caption{Different steps in the \texttt{FedDF}~\cite{feddf} algorithm.} 
    \label{fig:feddfworkflow}
      \shrinkspace
      \vspace{-5mm}
\end{figure}

The \texttt{FedDF}~\cite{feddf} algorithm proposes an ensemble distillation approach for model fusion that supports heterogeneous FL client model architectures. In contrast to \texttt{FedMD} (\S\ref{sec:femd}), it supports distillation with an unlabelled public dataset that can also be generated using Generative adversarial networks. As a result, it does not require any changes to the client training process. The different synchronous steps involved in the \texttt{FedDF} strategy are shown in Figure~\ref{fig:feddfworkflow}. At the start of each training round, the central server randomly selects a group of clients to participate in that round. Following this, the randomly selected clients train their local teacher models for a specified number of epochs (\circled{1}). After local training, each participating client sends the updated model weights to the central server (\circled{2}). For each unique client model architecture, the central server aggregates the received model weights to initialize a student server model. To distill knowledge to the initialized server model, each teacher client model is evaluated on mini-batches of the public dataset, and their generated logit outputs are used to train the student server model using a custom loss function~\cite{feddf}. This training process continues until a stable validation loss is reached for the student server model (\circled{3}). After the ensemble distillation process, we obtain distilled server models for each unique model architecture that contains collaborative knowledge from all the participating clients across all model architectures. The weights from these distilled server models are then distributed back to the clients based on their respective model architectures for the next training round (\circled{4}). Steps \circled{1}-\circled{4} are repeated until the desired accuracy is reached for each model architecture.

\vspace{-1mm}
\section{Related Work}
\label{sec:relatedwork}

\subsection{Serverless Federated Learning}
\label{sec:serverlessfl}
Using serverless computing technologies, particularly FaaS, for designing efficient systems for FL is a relatively new research direction. Existing works in this domain can be categorized into two groups: (i) systems that employ serverless functions exclusively in the \texttt{central server} ~\cite{jayaram2022lambda, jitfl, jayaramadaptive} and (ii) systems that leverage serverless functions in both entities of an FL system~\cite{flox, serverlessfl, fedless} (\S\ref{sec:intro}). In~\cite{jayaram2022lambda}, Jayaram et al. propose \(\lambda\)-FL, a serverless aggregation strategy for FL to improve fault tolerance and reduce resource wastage. The authors use serverless functions as aggregators to optimize the aggregation of model parameters in conventional \texttt{FedAvg}~\cite{mcmahan2017communication} over several steps. They implement their prototype using message queues, \texttt{Kafka}, and use \texttt{Ray}~\cite{ray} as the serverless platform. In~\cite{jitfl} and~\cite{jayaramadaptive}, the authors extend their previous strategy to enable adaptive and just-in-time aggregation of client model updates using serverless functions. In the second group, \emph{FedKeeper}~\cite{serverlessfl} was the first serverless FL system that enabled the training of Deep Neural Network (DNN) models using FL for clients distributed across a combination of heterogeneous FaaS platforms. However, it lacked crucial features required for practical FL systems, such as security and support for large DNN models. To address these drawbacks, \emph{FedLess}~\cite{fedless} was introduced as an evolution of \emph{FedKeeper} with multiple new enhancements. These include: (i) support for multiple open-source and commercial FaaS platforms, (ii) authentication/authorization of client functions using AWS Cognito, (iii) training of arbitrary homogeneous DNN models using the \texttt{Tensorflow} library, and (iv) the privacy-preserving FL training of models using Differential Privacy~\cite{Mothukuri2021}. In addition, \emph{FedLess} incorporates several optimizations for serverless environments, such as global namespace caching, running average model aggregation, and federated evaluation. A more recent work by Kotsehub et al.~\cite{flox} introduces \emph{Flox}, a system built on the \texttt{funcX}~\cite{chard2020funcx} serverless platform. \emph{Flox} aims to separate FL model training/inference from infrastructure management, providing users with a convenient way to deploy FL models on heterogeneous distributed compute endpoints. However, its tight integration with \texttt{funcX} restricts its compatibility with other open-source or commercial FaaS platforms, limiting its applicability and generality. As a result, in this paper, we use \emph{FedLess} as the serverless FL system.

\newcommand\YAMLcolonstyle{\color{red}\mdseries}
\newcommand\YAMLkeystyle{\color{black}\bfseries}
\newcommand\YAMLvaluestyle{\color{blue}\mdseries}

\makeatletter

\newcommand\language@yaml{yaml}

\expandafter\expandafter\expandafter\lstdefinelanguage
\expandafter{\language@yaml}
{
frame=single,
captionpos=b,
numbers=left,
xleftmargin=2.6em,
framexleftmargin=2.7em,
keywords={true,false,null,y,n},
keywordstyle=\color{darkgray}\bfseries,
basicstyle=\YAMLkeystyle\tiny,                                 
sensitive=false,
comment=[l]{\#},
morecomment=[s]{/*}{*/},
commentstyle=\color{purple}\ttfamily,
stringstyle=\YAMLvaluestyle\ttfamily,
moredelim=[l][\color{orange}]{\&},
moredelim=[l][\color{magenta}]{*},
moredelim=**[il][\YAMLcolonstyle{:}\YAMLvaluestyle]{:},   
morestring=[b]',
morestring=[b]",
literate =    {---}{{\ProcessThreeDashes}}3
{>}{{\textcolor{red}\textgreater}}1
{|}{{\textcolor{red}\textbar}}1
{\ -\ }{{\mdseries\ -\ }}3,
}

\lst@AddToHook{EveryLine}{\ifx\lst@language\language@yaml\YAMLkeystyle\fi}
\makeatother

\newcommand\ProcessThreeDashes{\llap{\color{cyan}\mdseries-{-}-}}

\begin{lstlisting}[float, floatplacement=t, language=yaml, label={yaml:modelconfig}, caption={Example model description representing a 3-layer CNN in \texttt{YAML}. Ellipses signify parameters that are omitted for brevity.}, belowskip=-1.5\baselineskip]
model:
  name: "my network"
  dataset:
    name: "mnist"
  input_shape: [32, 32, 3]
  block:
    layers: &ref_block
      - name: &lay_1_ref "layer_1"
        type: Conv2D
        input: [auto]
        params: [...]
      - name: &lay_2_ref "layer_2"
        type: Dense
        input: [*lay_2_ref]
        params:
          units: 10
          activation: "softmax"
  layers:
    - name: &layer_1 "layer_1"
      type: Conv2D
      input: [auto]
      params: [...]
    - name: &layer_2 "layer_2"
      type: "ReferenceBlock"
      input: [*layer_1]
      layers: *ref_block
  outputs: [*layer_1, *layer_2]
\end{lstlisting}
\vspace{-3mm}
\subsection{Model-agnostic Serverless FL}
\label{sec:modelagserverlessfl}
Existing research on model-agnostic KD is limited to conventional Infrastructure-as-a-Service (IaaS) based FL systems (\S\ref{sec:kdinfl}). In addition, the experiments performed for \texttt{FedMD}~\cite{fedmd} are simulated locally with only $10$ participating clients and overlook crucial infrastructure components. As a result, they fail to provide a comprehensive understanding of the algorithm performance in distributed settings, which involves factors such as system heterogeneity and stragglers~\cite{huba2022papaya}. Moreover, the performed experiments do not account for varying levels of data heterogeneity among FL clients. Similarly, for \texttt{FedDF}~\cite{feddf}, little emphasis has been placed on the distributed infrastructure, its optimization, and execution time in such settings, leaving these aspects as future work.
To the best of our knowledge, this work represents the first implementations of multiple KD algorithms within the serverless FL paradigm. Furthermore, we comprehensively analyze the performance and cost of the two algorithms in a distributed setting with 100 participating clients for multiple ML tasks.

\section{System Design}
\label{sec:sysdesign}

\subsection{Supporting Heterogeneous Client Model Architectures}
\label{sec:supporthetero}

\begin{figure}[t]
    \centering
    \includegraphics[width=0.9\columnwidth]{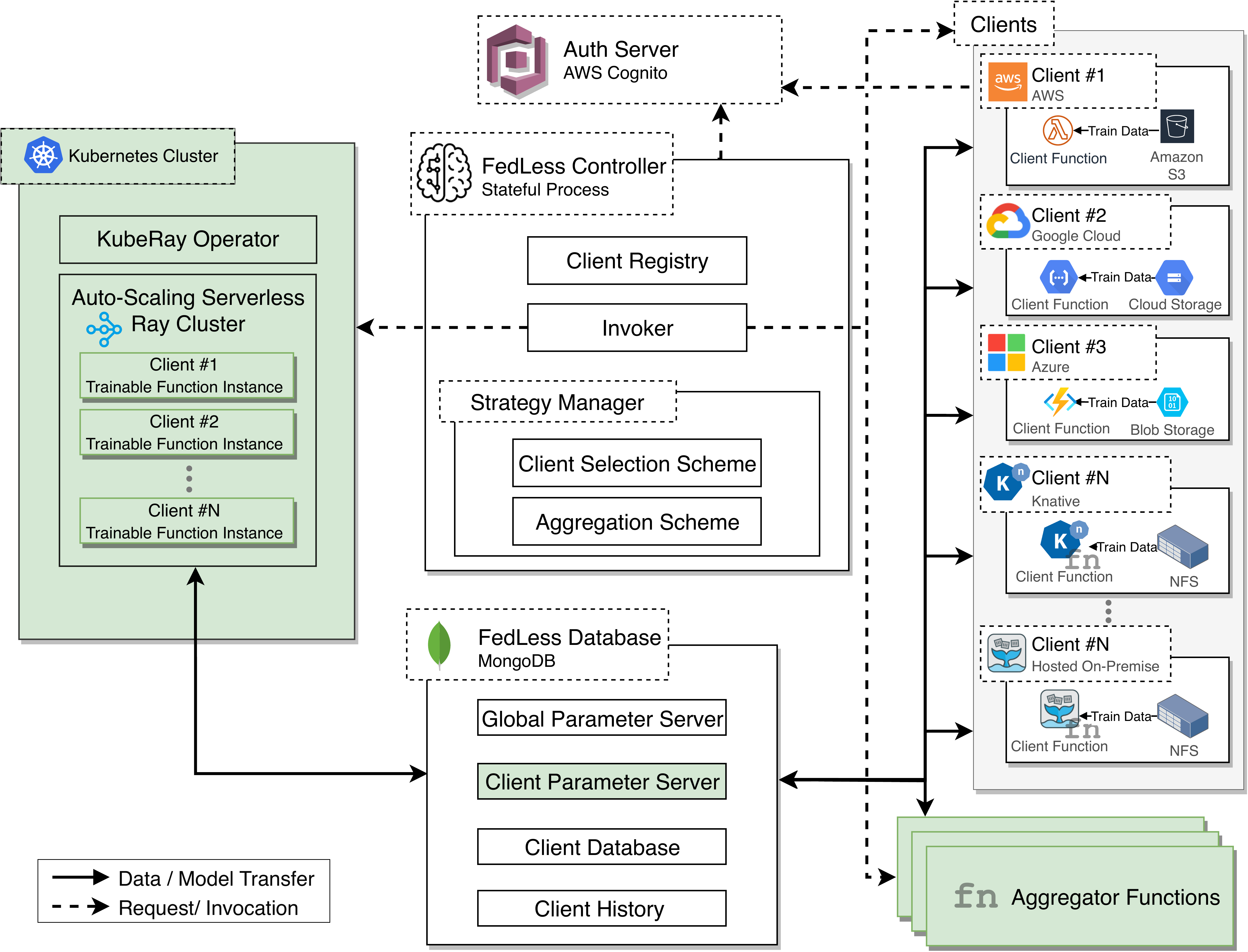}
    \caption{Proposed extensions (green) to \texttt{FedLess}~\cite{fedless}.} 
    \label{fig:fedless}
      \shrinkspace
\end{figure}

The initial version of \emph{FedLess}~\cite{fedless} only supports FL with homogenous client model architectures managed by the controller (Figure~\ref{fig:fedless}). To enable heterogeneous model training, we extend \emph{FedLess} to create and initialize diverse model architectures for each FL \texttt{client} based on the input configuration. Towards this, we implement a \texttt{Model Loader} module that parses a human-readable network configuration \texttt{YAML} file to generate and train arbitrary DNN models. 
Our module generates a directed acyclic graph (DAG) representing the DNN model from the input file that can be used directly to initialize a model in \texttt{TensorFlow}.

Listing~\ref{yaml:modelconfig} shows an example DNN model description written in \texttt{YAML}. Each element under the key \texttt{layers} describes a neural network unit where its inputs, parameters, and layer type are specified. Our module currently supports all layer types present till Keras version \texttt{2.7}. A \texttt{block} represents a repeatable collection of layers that can be defined once and re-used throughout the configuration file to define a network. To resolve dependencies between layers, our module relies on the key \texttt{name} for each layer. In addition, we use the anchor-alias (\texttt{\&,*}) syntax from the \texttt{YAML} standard to enable easy referencing and dereferencing of elements. For instance, the \texttt{layers} elements in the block are referred to as \texttt{\&ref\_block} in Line 7. After the file is parsed, those elements will appear as layers in the second layer (Line 26) due to the alias \texttt{*ref\_block}. However,  due to the dynamic nature of
referenceable blocks, defining inputs may lead to duplicating keys in the configuration file. To avoid this, our module uses the keyword \texttt{auto} to support automatic input interpretation. If a layer sets its input as \texttt{auto} then our module uses its parent to resolve its inputs. For instance, \texttt{\&lay\_1\_ref} (Line 8) will take its input from the input value described in Line 25. This relation between referenced blocks and the parent’s inputs enables nested blocks beyond the first-level nesting shown in Listing~\ref{yaml:modelconfig}. To this end, our implemented module offers data holders the flexibility to select any model architecture for their clients.

\subsection{Extending FedLess}
\label{sec:extfedless}

Figure~\ref{fig:fedless} presents an overview of the enhanced system architecture of \emph{FedLess}. The highlighted components in green represent our specific extensions to enable serverless KD with the system. To support unique model architectures for each FL client, we added a \emph{Client Parameter Server}. This parameter server is a \texttt{MongoDB} instance responsible for storing the model architecture and its hyperparameters for each client. To optimize certain steps in the workflow of the two individual algorithms, we made two major changes to the system. These changes are described in the following two aspects.

\subsubsection{Serverless Parallel Transfer Learning for \texttt{FedMD}}
\label{sec:serverparallelkd}
Prior to the distillation process, the \texttt{FedMD} algorithm requires an initial pre-training phase where all heterogeneous client models are trained on the public dataset until convergence  (\S\ref{sec:femd}). However, the time-intensive nature of this training process prevents its execution directly within the FaaS-based FL clients due to function time limitations imposed by most commercial FaaS platforms. For instance, AWS Lambda restricts the maximum function execution time to $15$ minutes~\cite{lambdaquotas}. Towards this, we extend \emph{FedLess} with the popular \texttt{Ray}~\cite{ray} distributed computing platform. We chose \texttt{Ray} due to its several advantages, such as arbitrarily long serverless functions, support for accelerators such as GPUs, seamless integration with popular Python libraries such as \texttt{numpy} and \texttt{Tensorflow}, and its elastic autoscaling capabilities. We deploy \texttt{Ray} on top of Kubernetes (k8s) using the \texttt{KubeRay} operator. In addition, we utilize the Ray tune~\cite{ray_tune} feature that enables seamless distributed parallel training of individual client models. The \emph{FedLess} controller can directly invoke client function instances for model training on the \texttt{Ray} cluster. On invocation, Ray creates \textit{actors} for each client function instance~\cite{ray}. These actors are independent Python processes that execute in a distributed and parallel manner across the cluster. 
Depending on the number of function invocations, Ray's autoscaler can request additional k8s pods to execute the client functions in parallel. Moreover, the autoscaler can also aggressively scale the number of pods to zero when resources are unused. By leveraging \texttt{Ray} in \emph{FedLess}, we can parallelize the initial transfer learning process using functions, support a large number of FL clients, and optimize resource utilization. To facilitate ease of use, we provide scripts to setup and configure \texttt{Ray} with \emph{FedLess}.

\subsubsection{Serverless Parallel Ensemble Distillation for \texttt{FedDF}}
\label{sec:parallelensembledist}

The \texttt{Fed\-DF} algorithm incorporates an ensemble distillation process, where knowledge is transferred from multiple teacher client models with heterogeneous architectures to a student server model  (\S\ref{sec:fedf}). In the original \texttt{FedDF} implementation~\cite{feddf}, this process is performed sequentially for each unique model architecture. However, since there are no dependencies across model architectures, this step can be easily parallelized. Towards this, we extend \emph{FedLess} to support multiple aggregator functions. Each aggregator function is responsible for performing the ensemble distillation process for a specific model architecture. These functions are triggered in parallel as soon as the distillation process begins. Moreover, these functions are generic and can be implemented using any FaaS platform. This design choice provides developers greater flexibility and prevents vendor/platform lock-in. To enable ease of development, we provide reference implementations of these functions for \texttt{OpenFaaS}~\cite{openfaas} and \texttt{Knative} platforms. By parallelizing ensemble distillation across model architectures in \emph{FedLess}, we can significantly decrease training times in practical FL systems, as demonstrated in our evaluation.

\begin{figure}[t]
    \centering
    \includegraphics[width=\columnwidth]{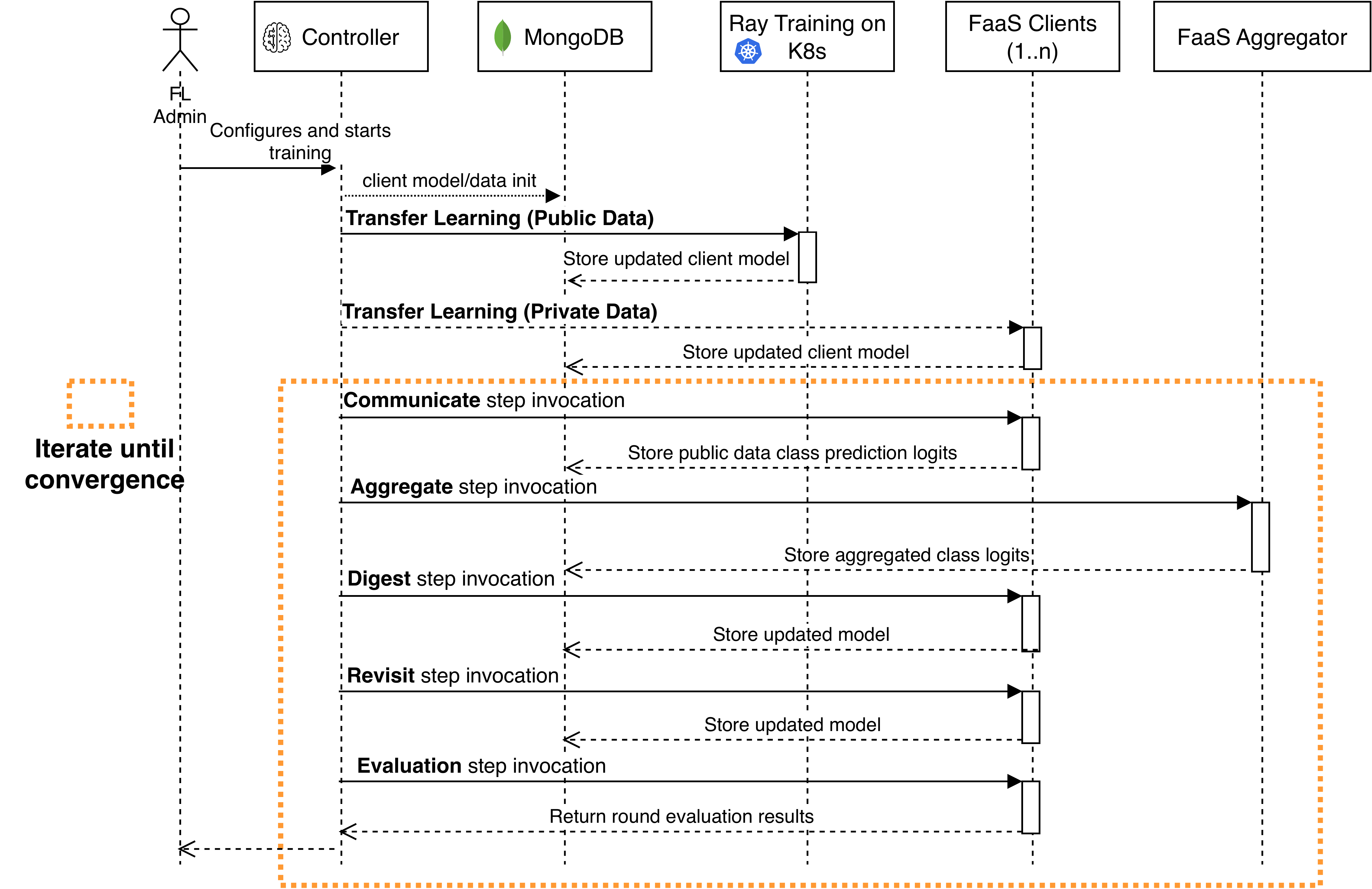}
    \caption{Serverless \texttt{FedMD} workflow.} 
    \label{fig:serverlessfedmdworkflow}
      \shrinkspace
      \vspace{-2mm}
\end{figure}

\section{Serverless Knowledge Distillation}
\label{sec:serverlesskd}
\subsection{Serverless FedMD}
\label{sec:severlessfedmd}
The training workflow for \texttt{FedMD} in the serverless paradigm with \emph{FedLess} is shown in Figure~\ref{fig:serverlessfedmdworkflow}. For simplicity, we omit some technical details related to the authentication/authorization of client functions, as well as some other minor interactions with the parameter server. Initially, the FL admin configures the client models, datasets, and the required client and \texttt{FedMD}-specific hyperparameters before starting the training process. Following this, the \emph{FedLess} controller (\S\ref{sec:sysdesign}) initializes the heterogeneous client models according to the provided configuration files (\S\ref{sec:supporthetero}) and creates the required data loaders for each client. In \texttt{FedMD}, each FL client has access to three datasets: (i) a private training dataset, (ii) a private testing dataset, and a (iii) public training dataset (\S\ref{sec:femd}). After this, we perform a one-time initial transfer learning process for all the individual client models before starting the collaboration phase (\S\ref{sec:femd}). To optimize this process, we divide it into two consecutive steps. In the initial step, we leverage \texttt{Ray} (\S\ref{sec:serverparallelkd}) to simultaneously train all client models on the public dataset until convergence. In the second step, the controller triggers individual FL clients to perform additional training of their models on their private datasets.  Following the initial transfer learning phase, we proceed to the collaborative training rounds, where the clients engage in knowledge distillation to collectively improve overall test accuracies. Each collaborative training round consists of five steps, which are executed as individual invocations of FL clients and the aggregator. In the communication step, all FL clients generate predictions on a random subset of the public dataset and store the corresponding prediction logits in \texttt{MongoDB}. Following this, the controller invokes the aggregator function, which aggregates the prediction logits from all the FL clients. In the digest step, each client trains its model to approach the aggregated prediction logits. The objective is to achieve logit alignment, thereby facilitating knowledge distillation among the clients. Following this, each client fine-tunes their model on their private dataset for a few epochs for model personalization. Finally, in the evaluation step, the controller performs an evaluation invocation to obtain the performance of each client model on the public test dataset. The five different steps continue until the desired accuracy is reached on the individual client models. 

\begin{figure}[t]
    \centering
    \includegraphics[width=\columnwidth]{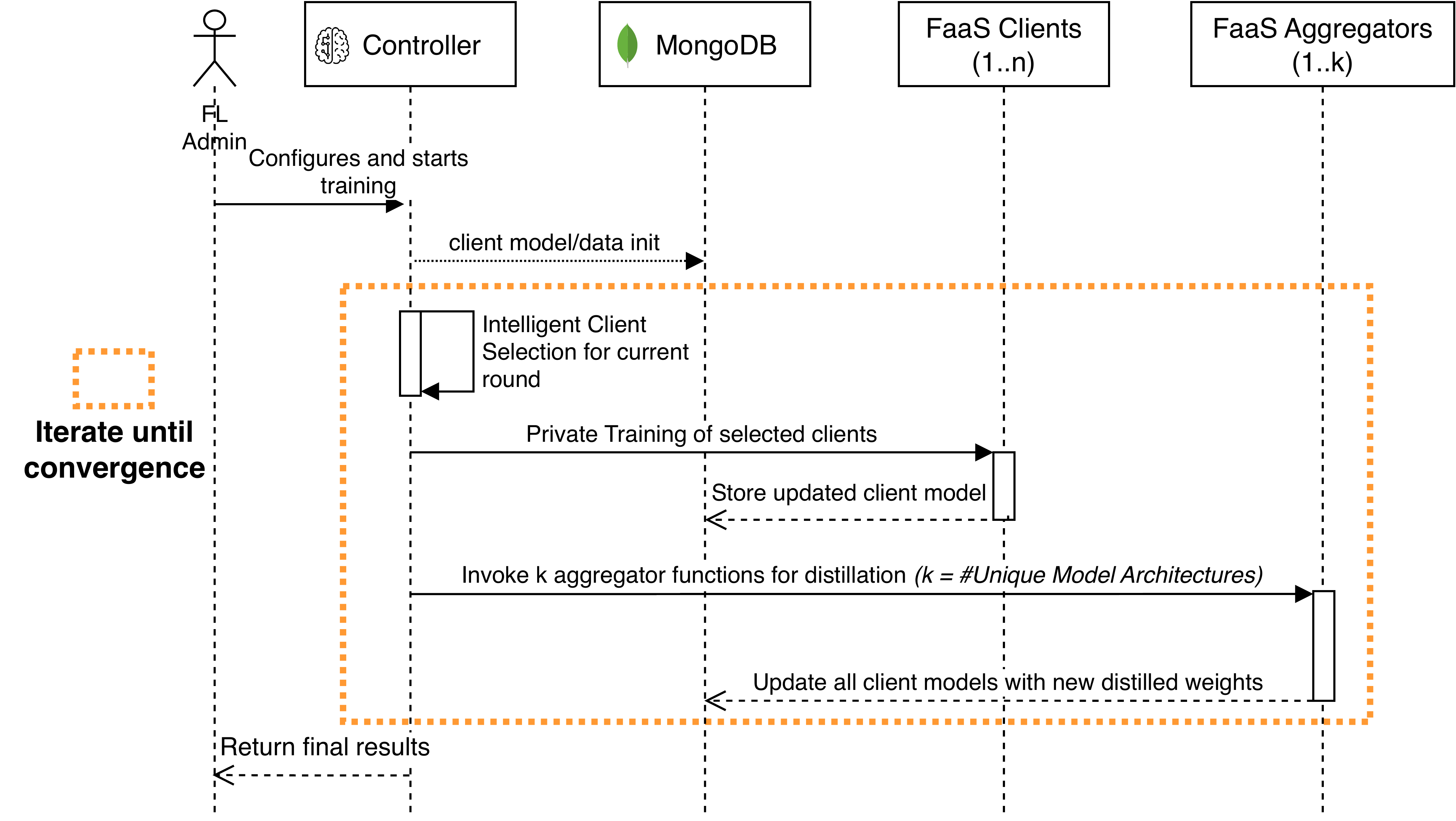}
    \caption{Serverless \texttt{FedDF} workflow.} 
    \label{fig:serverlessfeddfworkflow}
      \shrinkspace
      \vspace{-2mm}
\end{figure}

\vspace{-4mm}
\subsection{Serverless FedDF}
\label{sec:severlessfeddf}
Figure~\ref{fig:serverlessfeddfworkflow} shows the training workflow for the server-side distillation algorithm \texttt{FedDF} with \emph{FedLess}. Similar to \texttt{FedMD} (\ref{sec:severlessfedmd}), the FL admin first configures the client models, datasets, as well as the required client and \texttt{FedDF}-specific hyperparameters before starting the training process. After this, the iterative training process in \texttt{FedDF} is initiated. In the original \texttt{FedDF}~\cite{feddf} algorithm (\S\ref{sec:fedf}), a random subset of clients is selected to participate in each training round. However, due to resource and statistical heterogeneity of clients in FL, random selection can often lead to significantly higher training times because of \textit{stragglers}, i.e., slow clients~\cite{oort}. To mitigate this, we adapt the intelligent clustering-based client selection algorithm in \texttt{FedLesScan}~\cite{fedlesscan} and integrate it with \texttt{FedDF}. \texttt{FedLesScan} is a training strategy designed to facilitate efficient FL in serverless environments. It incorporates an adaptive clustering-based client selection algorithm that selects a subset of clients for
training based on their behavior in previous training rounds. However, the original clustering process in \texttt{FedLesScan} does not account for the heterogeneous client model architectures. This is important because clustering metrics such as training times can vary based on the complexity of different model architectures. In the extended version of \texttt{FedLesScan}, we first perform clustering among clients having the same model architecture. Following this, we perform a round-robin selection of the \textit{sorted clients}~\cite{fedlesscan} from each client model architecture group until the desired number of clients is selected. For sorting clients, we use the metric exponential moving average that relies on the training duration of clients.

In the initial step of the iterative training process, the controller uses our adapted \texttt{FedLesScan} algorithm to intelligently select a subset of clients from the client pool. Following this, the chosen clients are invoked to perform training on their private datasets. After training, the clients store the updated models back in the \texttt{MongoDB} database. Once all clients have finished training, the controller invokes the different aggregator functions for the ensemble distillation process (\S\ref{sec:parallelensembledist},\S\ref{sec:fedf}). These aggregator functions are invoked in parallel for each unique client model architecture. After completion, each aggregator function updates the weights of clients associated with their particular model architecture with the obtained new distilled weights. The iterative training process continues until a desired test accuracy is reached for each model architecture. 

\section{Experimental Setup}
\label{sec:expsetup}

\subsection{Datasets}
\label{sec:datasetclientdata}

In our experiments, we evaluate the implemented serverless KD algorithms (\S\ref{sec:serverlesskd}) for a variety of tasks. Table~\ref{table:experiment_datasets_fedmd} shows the different public/private datasets for the two algorithms. 

MNIST is a handwritten digit dataset comprising of $60,000$ training images and $10,000$ testing images. It has ten classes corresponding to the respective digits. On the other hand, EMNIST Letters consists of handwritten letters from the English language, comprising 145,600 characters distributed evenly among 26 classes. For more complex image classification tasks, we use the CIFAR-10 and CIFAR-100 datasets. Both datasets have $60,000$ color images divided uniformly into ten and $100$ mutually exclusive classes, respectively. For the language modeling domain, we use the Shakespeare dataset from the \texttt{LEAF}~\cite{caldas2018leaf} FL benchmark suite and the openly available Nietzsche text corpus~\cite{nietsche}. The Shakespeare dataset consists of $4,226,158$ sequences of length $80$ across $1,129$ different users, while the Nietzsche dataset consists of $200,271$ sequences of length $80$. In Shakespeare, each user has their own training and testing dataset. For both datasets, the task is to predict the next character given a sequence.

\begin{table}[t]
\centering
\begin{adjustbox}{width=0.9\columnwidth,  center}
\begin{tabular}{|c|c|c|cc|cc|}
\hline
\multirow{2}{*}{No.} & \multicolumn{1}{l|}{\multirow{2}{*}{Task}} & \multirow{2}{*}{Type} & \multicolumn{2}{c|}{FedMD}                              & \multicolumn{2}{c|}{FedDF}                            \\ \cline{4-7} 
                     & \multicolumn{1}{l|}{}                      &                       & \multicolumn{1}{c|}{Private}           & Public         & \multicolumn{1}{c|}{Private}     & Public (Unlabeled) \\ \hline
1                    & CV                                         & Classification        & \multicolumn{1}{c|}{MNIST}             & EMNIST Letters & \multicolumn{1}{c|}{MNIST}       & EMNIST Letters     \\ \hline
2                    & CV                                         & Classification        & \multicolumn{1}{c|}{CIFAR100 (Subset)} & CIFAR10        & \multicolumn{1}{c|}{CIFAR10}     & CIFAR100           \\ \hline
3                    & NLP                                        & Character Prediction  & \multicolumn{1}{c|}{Shakespeare}       & Nietzsche      & \multicolumn{1}{c|}{Shakespeare} & Nietzsche          \\ \hline
\end{tabular}

\end{adjustbox}
\caption{Evaluation tasks and datasets for serverless \texttt{FedMD} and \texttt{FedDF}.}
\shrinkspace
\vspace{-3mm}
\label{table:experiment_datasets_fedmd}
\end{table}

\begin{figure*}
 \begin{subfigure}{0.33\textwidth}
    \centering
        \includegraphics[width=0.65\columnwidth]{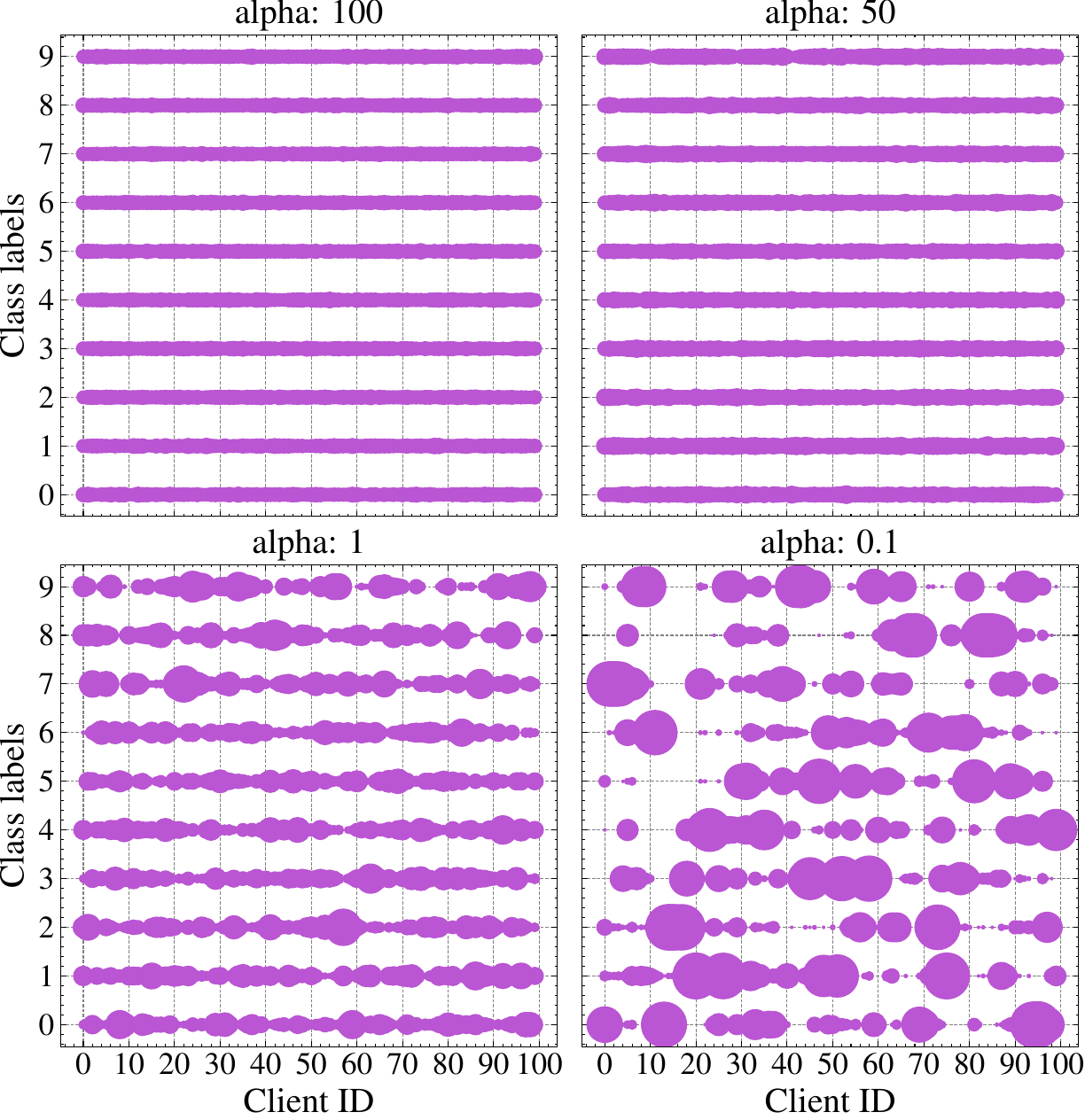}
        \caption{\texttt{FedMD}/\texttt{FedDF} with MNIST.}
        \label{fig:fedmdmnistdata}
\end{subfigure}
\begin{subfigure}{0.33\textwidth}
    \centering
        \includegraphics[width=0.65\columnwidth]{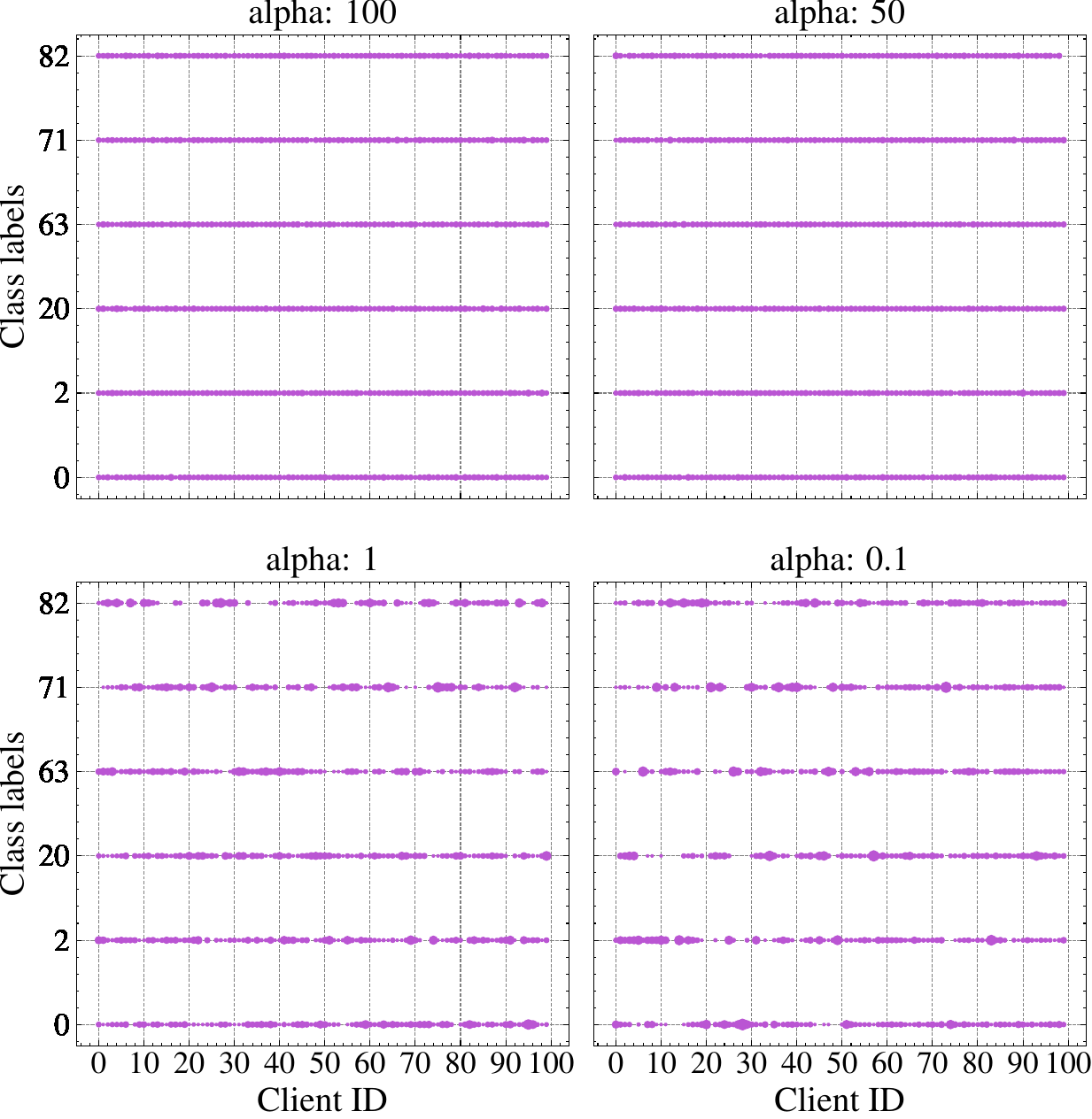}
        \caption{\texttt{FedMD} with CIFAR-100 (Subset).}
        \label{fig:fedmdcifar100}
\end{subfigure}
 \begin{subfigure}{0.33\textwidth}
    \centering
        \includegraphics[width=0.65\columnwidth]{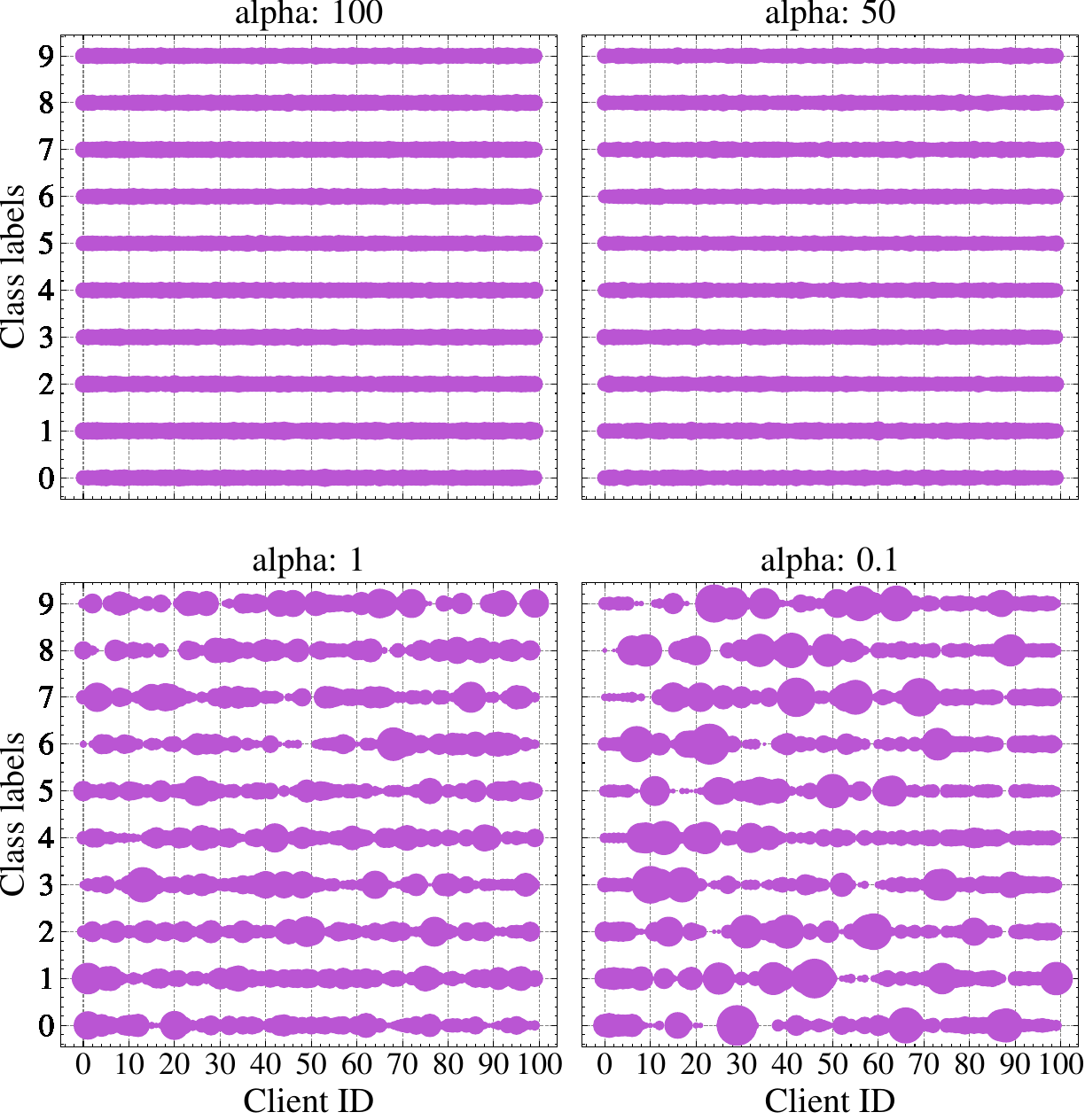}
        \caption{\texttt{FedDF} with CIFAR-10.}
        \label{fig:feddfcifar10}
\end{subfigure}
\caption{Private client training data distributions for the different algorithms and datasets for varying values of \(\alpha\). For a particular client ID, the size of each data point represents the number of samples belonging to a particular class.}
\label{fig:armresults} 
\shrinkspace
\end{figure*}

\subsection{Heterogeneous Client Data Distribution}
\label{sec:heterodata}
One of the major challenges in FL is the non-independent and identical (non-IID) data distribution among the participating clients~\cite{hsieh2020non} (\S\ref{sec:intro}). To analyze the behavior and robustness of the two serverless KD strategies toward different degrees of data heterogeneity, we use the Dirichlet distribution as in~\cite{hsu2019measuring} to create disjoint non-IID client training data partitions. A parameter \(\alpha\) controls the degree of non-IID data distribution, where a smaller \(\alpha\) value increases the probability of clients holding training samples from only one class and vice-versa. 

\begin{table}[t]
\centering
\begin{adjustbox}{width=0.9\columnwidth,  center}
\begin{tabular}{|c|c|c|c|c|c|}
\hline
\multicolumn{1}{|l|}{Model ID} & \multicolumn{1}{l|}{\#Clients} & \multicolumn{1}{l|}{\#Layer 1 Conv. Filters} & \multicolumn{1}{l|}{\#Layer 2 Conv. Filters} & \multicolumn{1}{l|}{\#Layer 3 Conv. Filters} & \multicolumn{1}{l|}{\#Trainable Parameters} \\ \hline
0                              & 10                             & 128                                          & 256                                          & -                                            & 729,856                                     \\ \hline
1                              & 30                             & 128                                          & 512                                          & -                                            & 1,458,176                                   \\ \hline
2                              & 20                             & 64                                           & 128                                          & 128                                          & 193,280                                     \\ \hline
3                              & 20                             & 64                                           & 128                                          & 256                                          & 352,640                                     \\ \hline
4                              & 10                             & 128                                          & 128                                          & 128                                          & 226,816                                     \\ \hline
5                              & 10                             & 128                                          & 128                                          & 256                                          & 386,176                                     \\ \hline
\end{tabular}
\end{adjustbox}
\caption{Different client CNN models for \texttt{FedMD}/\texttt{FedDF} in task one (Table~\ref{table:experiment_datasets_fedmd}).}
\shrinkspace  
\vspace{-4mm}
\label{table:client_model_fedmd_mnist}
\end{table}

\begin{table}[t]
\centering
\begin{adjustbox}{width=0.9\columnwidth,  center}
\begin{tabular}{|c|c|c|c|c|c|}
\hline
\multicolumn{1}{|l|}{Model ID} & \multicolumn{1}{l|}{\#Clients} & \multicolumn{1}{l|}{\#Layer 1 Conv. Filters} & \multicolumn{1}{l|}{\#Layer 2 Conv. Filters} & \multicolumn{1}{l|}{\#Layer 3 Conv. Filters} & \multicolumn{1}{l|}{\#Trainable Parameters} \\ \hline
0                              & 10                             & 128                                          & 256                                          & -                                            & 729,856                                     \\ \hline
1                              & 10                             & 64                                           & 128                                          & 192                                          & 274,112                                     \\ \hline
2                              & 30                             & 64                                           & 64                                           & 128                                          & 104,128                                     \\ \hline
3                              & 30                             & 64                                           & 128                                          & 256                                          & 352,640                                     \\ \hline
4                              & 20                             & 128                                          & 128                                          & 128                                          & 226,816                                     \\ \hline
\end{tabular}
\end{adjustbox}
\caption{Different client CNN models for \texttt{FedMD}/\texttt{FedDF} in task two (Table~\ref{table:experiment_datasets_fedmd}).}
\shrinkspace
\vspace{-5mm}
\label{table:client_model_fedmd_cifar}
\end{table}

Figure~\ref{fig:fedmdmnistdata} visualizes the private client training data distributions for the MNIST dataset for $100$ clients and different values of \(\alpha\). A value of \(\alpha=100\) represents uniform data distribution, while a value of \(\alpha=0.1\) represents an extreme non-IID scenario. For task one (Table~\ref{table:experiment_datasets_fedmd}), we  use the same public and private data distributions for \texttt{FedMD} and \texttt{FedDF}. However, in \texttt{FedDF}, only the image features and no labels are utilized for the distillation process. Each client model gets evaluated on the complete MNIST test dataset comprising $10000$ testing images (\S\ref{sec:datasetclientdata}). In task two for \texttt{FedMD}, we use a subset of six classes from the CIFAR-100 dataset as in~\cite{fedmd} for the main learning task. Figure~\ref{fig:fedmdcifar100} shows the private client data distribution in this scenario for various values of \(\alpha\). The complete CIFAR-10 dataset is employed as the public distillation dataset for this task (\S\ref{sec:femd}). During testing, each client's performance is evaluated on the complete global CIFAR-100 test dataset subsetted for the six classes. On the other hand, for \texttt{FedDF} in task two, we use CIFAR-10 as the private dataset and CIFAR-100 as the public dataset. The private training data distribution for \texttt{FedDF} with CIFAR-10 is shown in Figure~\ref{fig:feddfcifar10}. For evaluating individual client models, we use the complete CIFAR-10 test dataset comprising $10000$ images. Similar to task one, we use the same public and private dataset distributions for the two algorithms in task three, as shown in Table~\ref{table:experiment_datasets_fedmd}. For the non-IID scenario, we use the pre-provided non-IID partitions for the Shakespeare dataset from the \texttt{LEAF} FL benchmark suite~\cite{caldas2018leaf}. We don't use the Dirichlet distribution for creating non-IID partitions for Shakespeare  since that is only suitable for classification tasks. Note that for all tasks, the public dataset is uniformly distributed.

\subsection{Heterogeneous Client Model Architectures}
\label{sec:clientmodels}

In our experiments, we use multiple model architectures distributed among 100 participating clients based on the machine learning task (Table~\ref{table:experiment_datasets_fedmd}). For tasks one and two, we utilize 2-layer and 3-layer convolutional neural networks (CNNs). These model architectures are unevenly distributed among the clients to simulate real-world scenarios, as shown in Tables~\ref{table:client_model_fedmd_mnist} and~\ref{table:client_model_fedmd_cifar}. Each convolution filter layer is followed by batch normalization, ReLU activation, and a dropout of $0.2$. Dropout enables regularization and prevents client models from overfitting on small amounts of private local data. To ensure a fair comparison between the performance of \texttt{FedMD} and \texttt{FedDF} in the serverless paradigm, we use the same client model distributions for both algorithms. For task three, we utilize LSTM Recurrent Neural Networks with a single layer and a varying number of units, as shown in Table~\ref{table:client_model_fedmd_shakespeare}. Every network takes an input sequence length of 80, followed by an initial embedding size of 8 and then a single LSTM layer. For all tasks, we chose the model architectures that have been used previously in this domain~\cite{fedmd, fedlesscan, feddf, serverlessfl}. Due to space limitations, we omit the specific hyperparameters used for the two algorithms but will describe them in detail in our code repository.

\begin{table}[t]
\centering
\begin{adjustbox}{width=0.7\columnwidth,  center}
\begin{tabular}{|c|c|c|c|c|}
\hline
\multicolumn{1}{|l|}{Model ID} & \multicolumn{1}{l|}{\#Clients} & \multicolumn{1}{l|}{\#Units} & \multicolumn{1}{l|}{Embedding Dim} & \multicolumn{1}{l|}{\#Trainable Parameters} \\ \hline
0                              & 60                             & 128                          & 8                                  & 81,378                                      \\ \hline
1                              & 10                             & 64                           & 8                                  & 24,674                                      \\ \hline
2                              & 30                             & 256                          & 8                                  & 293,090                                     \\ \hline
\end{tabular}
\end{adjustbox}
\caption{Different client LSTM models for \texttt{FedMD}/\texttt{FedDF} in task three (Table~\ref{table:experiment_datasets_fedmd}).}
\label{table:client_model_fedmd_shakespeare}
\shrinkspace
\vspace{-5mm}
\end{table}

\begin{figure*}[t]

 \begin{subfigure}{0.195\textwidth}
    \centering
        \includegraphics[width=\columnwidth]{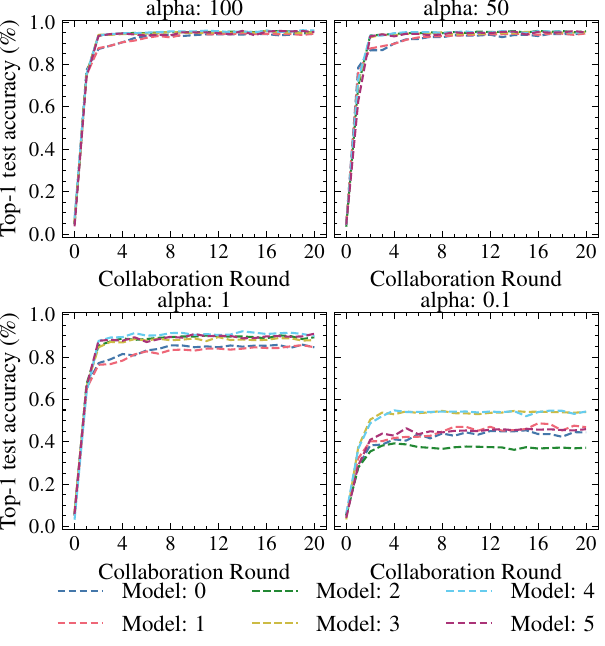}
        \caption{\texttt{FedMD} with MNIST.}
        \label{fig:fedmdmnist}
\end{subfigure}
\begin{subfigure}{0.195\textwidth}
    \centering
        \includegraphics[width=\columnwidth]{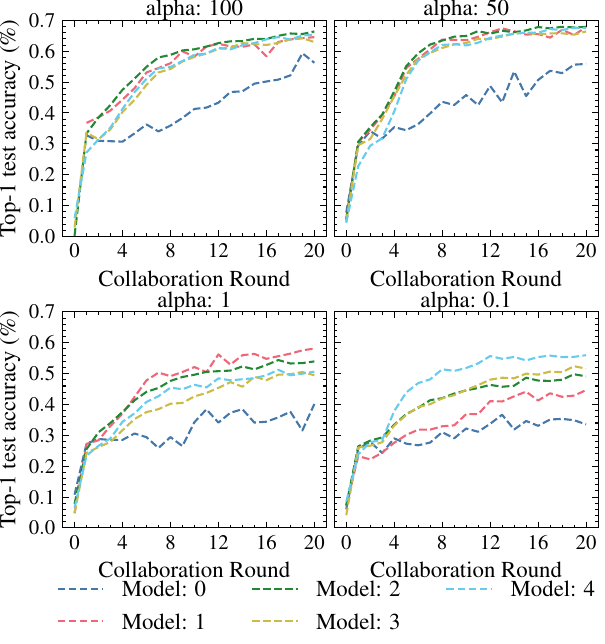}
        \caption{\texttt{FedMD} with CIFAR.}
        \label{fig:fedmdcifar}
\end{subfigure}
 \begin{subfigure}{0.195\textwidth}
    \centering
        \includegraphics[width=\columnwidth]{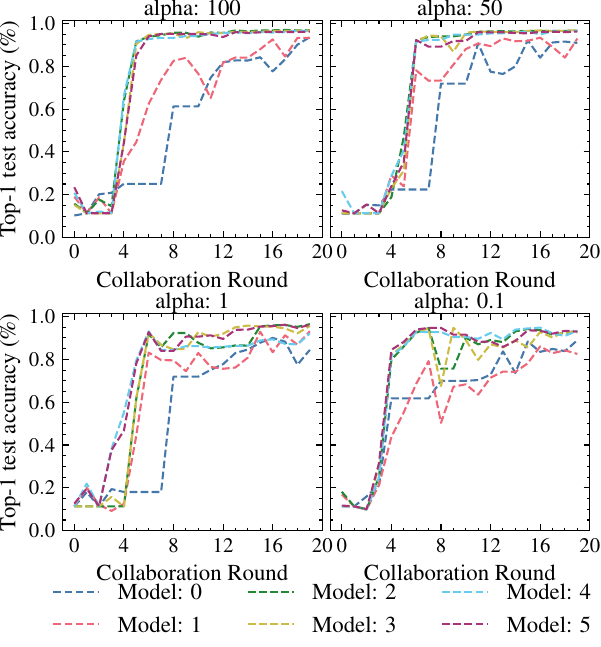}
        \caption{\texttt{FedDF} with MNIST.}
        \label{fig:feddfmnist}
\end{subfigure}
 \begin{subfigure}{0.195\textwidth}
    \centering
        \includegraphics[width=\columnwidth]{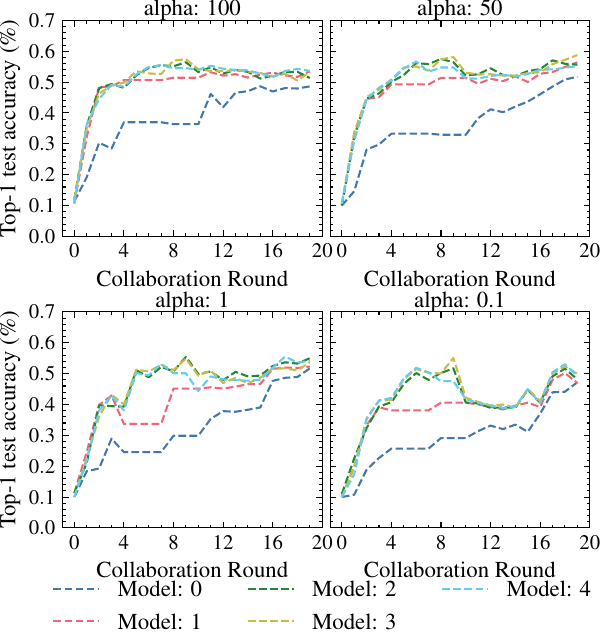}
         \caption{\texttt{FedDF} with CIFAR.}
         \label{fig:feddfcifar}
\end{subfigure}
 \begin{subfigure}{0.195\textwidth}
    \centering
        \includegraphics[width=\columnwidth]{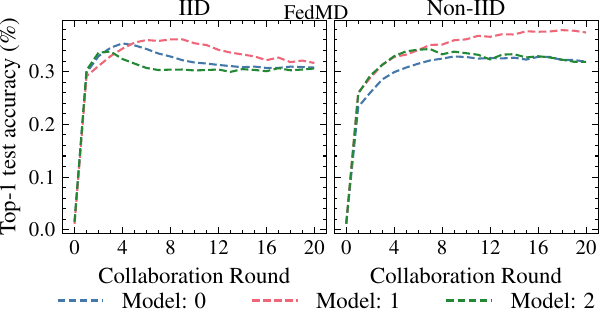}
         \includegraphics[width=\columnwidth]{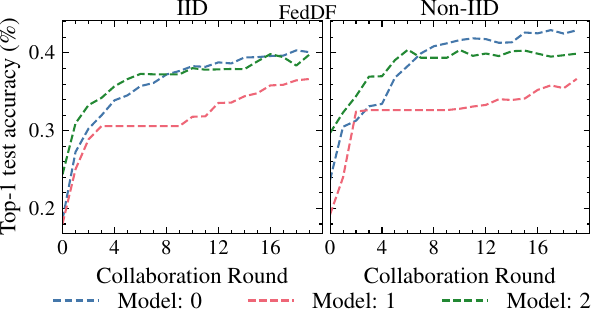}
         \caption{Shakespeare dataset.}
         \label{fig:fedmdshakes}
\end{subfigure}

\vspace{-2mm}
\caption{Comparing Top-1 model accuracies across heterogeneous model architectures for the serverless implementations of \texttt{FedMD} and \texttt{FedDF}.}
\label{fig:accresultallmmodel} 
\shrinkspace
\end{figure*}

\vspace{-2mm}
\subsection{Infrastructure Setup}
\label{sec:infrasetup}

For our experiments, we deployed \emph{FedLess} (\S\ref{sec:extfedless}) on a virtual machine hosted on our institute's compute cloud. The VM was configured with $40$vCPUs and $177$GiB of RAM. For the training cluster utilizing \texttt{Ray} (\S\ref{sec:serverparallelkd}), we assigned each function instance with 4 vCPUs and 16 GiB of RAM. Moreover, we set the maximum number of function instances to eight in the 
\texttt{Ray} autoscaler. Furthermore, for hosting our datasets, we used a nginx store running on a VM with 10vCPUs and 45GiB of RAM.


For FaaS-based FL clients and the different aggregator functions, we used OpenFaaS~\cite{openfaas} as the FaaS platform. We used $100$ client functions for our experiments. Each client had a limit of $2$vCPUs and $4$GiB of RAM. For \texttt{FedMD}, we sample $100$ clients per round  while limiting the number of clients per round to ten for \texttt{FedDF}. We deployed a single aggregator function for \texttt{FedMD} with $4$vCPUs and $8$GiB of RAM. In contrast, for \texttt{FeDF}, we used six aggregator functions (\S\ref{sec:parallelensembledist}). Each aggregator function had a limit of $6$vCPUs and $16$GiB of RAM.

\vspace{-3mm}

\section{Experimental Results}
\label{sec:expresults}
For all our experiments, we follow best practices while reporting results and repeat them three times.

\vspace{-3mm}
\subsection{Comparing Accuracy}
\label{sec:compacc}

In this subsection, we focus on demonstrating the convergence and improved accuracy obtained through serverless knowledge distillation among heterogeneous client models rather than pursuing state-of-the-art accuracies on these tasks. Towards this, we limit our experiments to $20$ communication rounds for the two algorithms across all datasets and levels of data heterogeneity (\S\ref{sec:datasetclientdata},\S\ref{sec:heterodata}). Figure~\ref{fig:accresultallmmodel} presents the top-1 model accuracies for \texttt{FedMD} and \texttt{FedDF} across the different datasets. Top-1 model accuracy is a common evaluation metric and represents the accuracy of a model in correctly predicting the most probable class label for a given input out of all possible class labels. For \texttt{FedMD}, each line in Figures~\ref{fig:fedmdmnist},~\ref{fig:fedmdcifar}, and~\ref{fig:fedmdshakes} represents the average model accuracy across all clients belonging to a particular unique model architecture group (\S\ref{sec:clientmodels}). On the other hand, for \texttt{FedDF}, each line in Figures~\ref{fig:feddfmnist},~\ref{fig:feddfcifar}, and~\ref{fig:fedmdshakes} corresponds to the accuracy of the global server model for that particular unique model architecture group. For the MNIST dataset with \texttt{FedMD}, we observe that the KD process is predominantly concentrated in the initial collaboration rounds. Following this, the accuracy curve stabilizes, indicating a plateau in performance improvement. For the scenario with uniform data distribution (\S\ref{sec:heterodata}), we observed a maximum accuracy of $96$\% for model 4 as shown in Figure~\ref{fig:fedmdmnist}. In contrast, for \(\alpha=0.1\), we observed a $46.5$\% average accuracy drop across all model architectures as compared to the IID scenario with \(\alpha=100\). This can be attributed to the divergence of the globally aggregated logits due to the significantly high variance in the private data distribution among the FL clients~\cite{hsieh2020non}(\S\ref{sec:femd},\S\ref{sec:severlessfedmd}). In contrast to the MNIST dataset, we observe a more gradual increase in accuracy with collaboration rounds for the CIFAR dataset with \texttt{FedMD}. For the i.i.d. scenario, we observe a maximum test accuracy of $66.4$\% with model two as shown in Figure~\ref{fig:fedmdcifar}. Furthermore, we observe a $13$\% average drop in accuracy for \(\alpha=0.1\) as compared to the scenario with uniform data distribution. Figure~\ref{fig:fedmdshakes} shows the performance of \texttt{FedMD} on the Shakespeare dataset for the IID and non-IID scenarios. For both scenarios, model one achieved the maximum accuracy of $36$\% and $37.8$\%, respectively. We observe faster convergence for the different model architectures in the IID scenario but do not observe any significant difference in the highest achieved accuracy between the two scenarios. This is because in \texttt{FedMD}, the transfer learning step was performed on the Nietzsche dataset (\S\ref{sec:datasetclientdata}) that predicts the same set of classes as the  Shakespeare dataset. Hence, the initial models after the transfer learning process were already strong in character prediction, and the subsequent collaborative fine-tuning among the FL clients only involved adapting the models to the specific characteristics of the Shakespeare dataset. Figure~\ref{fig:feddfmnist} shows the performance of the \texttt{FedDF} algorithm with the MNIST dataset for varying levels of data heterogeneity. We observe a maximum accuracy of $97$\% with \(\alpha=100\) and a accuracy of $94.8$\% with \(\alpha=0.1\). Figures~\ref{fig:feddfcifar} and~\ref{fig:fedmdshakes} show the performance of the \texttt{FedDF} algorithm with the CIFAR and the Shakespeare datasets, respectively. In the IID scenario, we observed maximum model accuracies of $57.5$\% and $40$\% for the two datasets. On the other hand, in the non-IID scenario, we observed maximum accuracies of $55$\% and $43$\% for the two datasets. We observe that \texttt{FedDF} demonstrates greater robustness to higher non-IID data distributions compared to \texttt{FedMD}, as it maintains higher model accuracies even for lower values of \(\alpha\). This can be attributed to two reasons. First, the usage of ensemble distillation in \texttt{FedDF} enhances the robustness of the global model to noise and outliers. Second, \texttt{FedDF} leverages unlabeled data in the distillation process, thereby enhancing its capacity to generalize to unseen data.



\subsection{Comparing Performance and Cost}
\label{sec:comperf}

\begin{figure*}

 \begin{subfigure}{0.33\textwidth}
    \centering
        \includegraphics[width=\columnwidth]{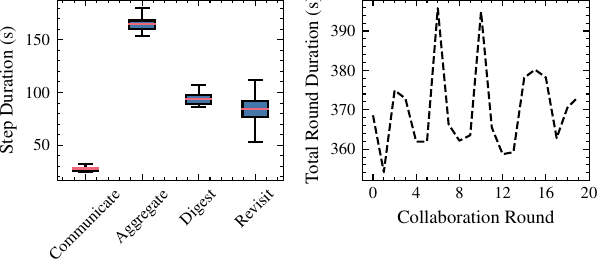}
        \caption{\texttt{FedMD} with MNIST.}
        \label{fig:fedmdmnisttiming}
\end{subfigure}
\begin{subfigure}{0.33\textwidth}
    \centering
        \includegraphics[width=\columnwidth]{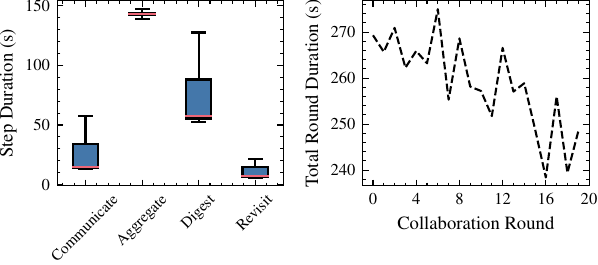}
        \caption{\texttt{FedMD} with CIFAR.}
        \label{fig:fedmdcifartiming}
\end{subfigure}
 \begin{subfigure}{0.33\textwidth}
    \centering
        \includegraphics[width=\columnwidth]{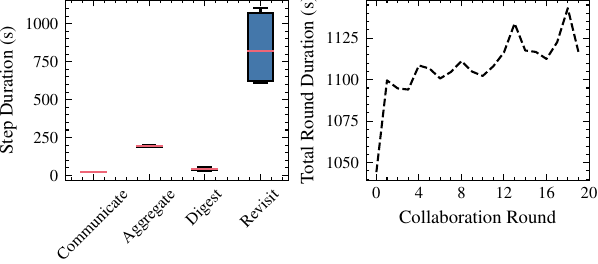}
        \caption{\texttt{FedMD} with Shakespeare.}
        \label{fig:fedmdshakestiming}
\end{subfigure}
 \begin{subfigure}{0.33\textwidth}
    \centering
        \includegraphics[width=\columnwidth]{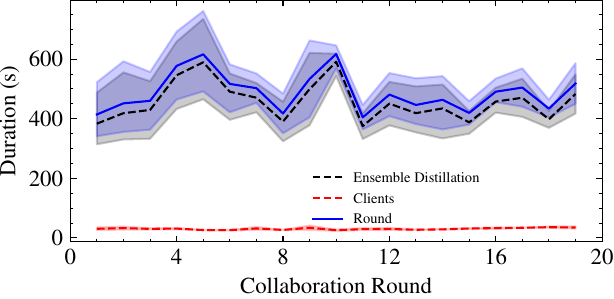}
        \caption{\texttt{FedDF} with MNIST.}
        \label{fig:feddfmnisttiming}
\end{subfigure}
 \begin{subfigure}{0.33\textwidth}
    \centering
        \includegraphics[width=\columnwidth]{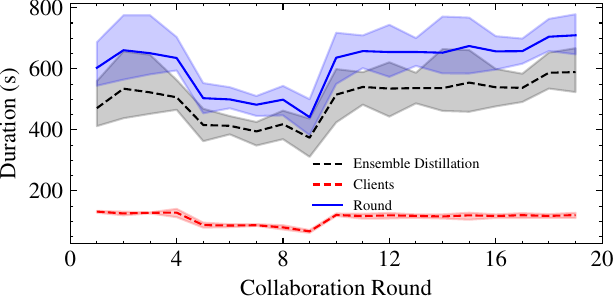}
        \caption{\texttt{FedDF} with CIFAR.}
        \label{fig:feddfcifartiming}
\end{subfigure}
 \begin{subfigure}{0.33\textwidth}
    \centering
        \includegraphics[width=\columnwidth]{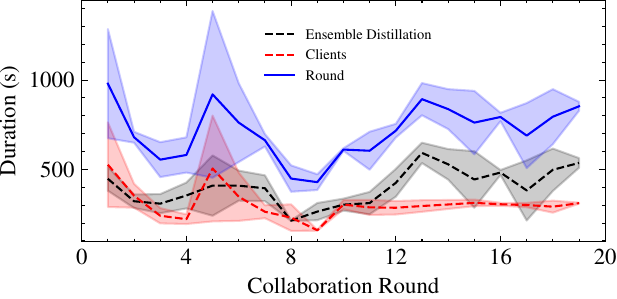}
        \caption{\texttt{FedDF} with Shakespeare.}
        \label{fig:feddfshakestiming}
\end{subfigure}
\caption{Comparing individual training steps and collaborative round durations for the serverless implementations of \texttt{FedMD} and \texttt{FedDF} across different datasets (\S\ref{sec:datasetclientdata}).}
\label{fig:timeresultallmmodel} 
\shrinkspace
\end{figure*}

Analyzing FL systems involves considering important factors such as the time and the cost required to complete different training tasks. This is particularly relevant in our case, as we utilize FaaS, in which users are only billed for the execution time of functions~\cite{rise}. To present summarized results, we aggregate the timings and costs across all clients for the different levels of data heterogeneity. For computing training costs, we use the cost computation model~\cite{gcf_cost} used by Google to estimate the cost for each function based on the number of invocations, allocated memory, and execution duration (\S\ref{sec:infrasetup}).

\begin{table}[t]
\centering
\begin{adjustbox}{width=0.9\columnwidth,  center}
\begin{tabular}{|c|c|c|c|c|c|c|c|}
\hline
Dataset                      & Metric         & Aggregate & Communicate & Revisit & \begin{tabular}[c]{@{}c@{}}Transfer Learning \\ (Private)\end{tabular} & Digest & Overall         \\ \hline
\multirow{2}{*}{MNIST}       & Duration (min) & 55.2      & 378.4       & 867     & 42.5                                                                   & 1329.8 & \textbf{2672.9} \\ \cline{2-8} 
                             & Cost (USD)      & 0.37      & 1.32        & 3.02    & 0.15                                                                   & 4.63   & \textbf{9.49}   \\ \hline
\multirow{2}{*}{CIFAR}       & Duration (min) & 47.7      & 579.4       & 111.6   & 6.5                                                                    & 746.4  & \textbf{1491.6} \\ \cline{2-8} 
                             & Cost (USD)      & 0.32      & 2.02        & 0.39    & 0.02                                                                   & 2.6    & \textbf{5.35}   \\ \hline
\multirow{2}{*}{Shakespeare} & Duration (min) & 64.5      & 300.8       & 5945.2  & 122.7                                                                  & 656.3  & \textbf{7089.5} \\ \cline{2-8} 
                             & Cost (USD)      & 0.43      & 1.05        & 20.69   & 0.43                                                                   & 2.28   & \textbf{24.88}  \\ \hline
\end{tabular}
\end{adjustbox}
\caption{Comparing total execution time and cost for serverless \texttt{FedMD} across the different datasets.}
\shrinkspace
\vspace{-4mm}
\label{table:fedmdserverlesscost}
\end{table}

\begin{table}[t]
\centering
\begin{adjustbox}{width=0.6\columnwidth,  center}
\begin{tabular}{|c|c|c|c|c|}
\hline
Dataset                      & Metric         & Aggregators & Clients & Overall         \\ \hline
\multirow{2}{*}{MNIST}       & Duration (min) & 152         & 56.4    & \textbf{208.4}  \\ \cline{2-5} 
                             & Cost (USD)      & 6.07        & 0.2     & \textbf{6.27}   \\ \hline
\multirow{2}{*}{CIFAR}       & Duration (min) & 163.87      & 213.55  & \textbf{377.42} \\ \cline{2-5} 
                             & Cost (USD)      & 5.46        & 0.74    & \textbf{6.2}    \\ \hline
\multirow{2}{*}{Shakespeare} & Duration (min) & 134.4       & 466.9   & \textbf{601.3}  \\ \cline{2-5} 
                             & Cost (USD)      & 2.68        & 1.62    & \textbf{4.3}    \\ \hline
\end{tabular}
\end{adjustbox}
\caption{Comparing total execution time and cost for serverless \texttt{FedDF} across the different datasets.}
\shrinkspace
\label{table:feddfserverlesscost}
\vspace{-4mm}
\end{table}

To offer detailed insights into the serverless KD training process, Figure~\ref{fig:timeresultallmmodel} shows the timings of the individual training steps and the collaborative round durations for the two algorithms. For relevance, we omit the timings for the initial one-time pre-training process in \texttt{FedMD} (\S\ref{sec:severlessfedmd}). For the MNIST dataset with \texttt{FedMD}, we observe that each collaborative training round takes between $360$ to $390$ seconds, as shown in Figure~\ref{fig:fedmdmnisttiming}. A significant portion of this time ($40$\%) is spent within the single aggregator function for aggregating prediction logits from all $100$ participating clients. The digest and revisit steps take a comparable amount of time, while the communicate step is the fastest as it involves only a forward pass inference on the public dataset by all clients (\S\ref{sec:severlessfedmd}). We observe relatively shorter round durations for the CIFAR dataset with \texttt{FedMD} compared to MNIST, as shown in Figure~\ref{fig:fedmdcifartiming}. This can be attributed to the shorter revisit step due to the smaller private client dataset as described in \S\ref{sec:heterodata}. For the Shakespeare dataset with \texttt{FedMD}, a majority of the time ($90$\%) is spent in the revisit step, as shown in Figure~\ref{fig:fedmdshakestiming}. This can be attributed to the significantly large number of epochs required by the LSTM text models for this training step as compared to the CNN models for image tasks (\S\ref{sec:clientmodels}). Figures~\ref{fig:feddfmnisttiming},~\ref{fig:feddfcifartiming}, and~\ref{fig:feddfshakestiming} present the total collaborative round duration along with the timings for private client training and ensemble distillation for the different datasets with \texttt{FedDF}. The highlighted area in these plots represents the variability observed for multiple runs across different data heterogeneity levels (\S\ref{sec:heterodata}). For the MNIST and CIFAR datasets, we observe that the majority of the round duration (\(>\)$90$\%) is spent in the ensemble distillation process that occurs in the aggregator functions for each unique model architecture (\S\ref{sec:severlessfeddf}). On the other hand, for the Shakespeare dataset, we observe similar durations for the ensemble distillation and private client training process. This can again be attributed to the higher number of local epochs required in private client training for LSTM networks.

\begin{table}[t]
\centering
\begin{adjustbox}{width=0.6\columnwidth,  center}
\begin{tabular}{|c|cccccc|}
\hline
\multirow{3}{*}{Model ID} & \multicolumn{6}{c|}{Maximum Top-1 accuracy (\%)}                                                                                                                                               \\ \cline{2-7} 
                          & \multicolumn{2}{c|}{MNIST \(\alpha=1\)}                                      & \multicolumn{2}{c|}{CIFAR \(\alpha=1\)}                                              & \multicolumn{2}{c|}{Shakespeare (non-IID)}                   \\ \cline{2-7} 
                          & \multicolumn{1}{c|}{FedMD} & \multicolumn{1}{c|}{FedDF}         & \multicolumn{1}{c|}{FedMD}         & \multicolumn{1}{c|}{FedDF}         & \multicolumn{1}{c|}{FedMD}         & FedDF         \\ \hline
0                         & \multicolumn{1}{c|}{0.86}  & \multicolumn{1}{c|}{\textbf{0.90}} & \multicolumn{1}{c|}{0.40}          & \multicolumn{1}{c|}{\textbf{0.52}} & \multicolumn{1}{c|}{0.33}          & \textbf{0.43} \\ \hline
1                         & \multicolumn{1}{c|}{0.86}  & \multicolumn{1}{c|}{\textbf{0.93}} & \multicolumn{1}{c|}{\textbf{0.58}} & \multicolumn{1}{c|}{0.52}          & \multicolumn{1}{c|}{\textbf{0.38}} & 0.37          \\ \hline
2                         & \multicolumn{1}{c|}{0.90}  & \multicolumn{1}{c|}{\textbf{0.96}} & \multicolumn{1}{c|}{0.54}          & \multicolumn{1}{c|}{\textbf{0.55}} & \multicolumn{1}{c|}{0.34}          & \textbf{0.40} \\ \hline
3                         & \multicolumn{1}{c|}{0.89}  & \multicolumn{1}{c|}{\textbf{0.96}} & \multicolumn{1}{c|}{0.51}          & \multicolumn{1}{c|}{\textbf{0.55}} & \multicolumn{1}{c|}{-}             & -             \\ \hline
4                         & \multicolumn{1}{c|}{0.92}  & \multicolumn{1}{c|}{\textbf{0.93}} & \multicolumn{1}{c|}{0.51}          & \multicolumn{1}{c|}{\textbf{0.56}} & \multicolumn{1}{c|}{-}             & -             \\ \hline
5                         & \multicolumn{1}{c|}{0.91}  & \multicolumn{1}{c|}{\textbf{0.96}} & \multicolumn{1}{c|}{-}             & \multicolumn{1}{c|}{-}             & \multicolumn{1}{c|}{-}             & -             \\ \hline
\end{tabular}
\end{adjustbox}
\caption{Comparing maximum Top-1 accuracy for \texttt{FedMD} and \texttt{FedDF} across all datasets and heterogeneous client model architectures.}
\label{table:comparision_max_accuracy}
\shrinkspace
\vspace{-5mm}
\end{table}

\begin{table}[t]
\centering
\begin{adjustbox}{width=0.7\columnwidth,  center}
\begin{tabular}{|c|c|c|c|}
\hline
Optimizations                  & MNIST & CIFAR & Shakespeare \\ \hline
Transfer Learning with \texttt{Ray}     & 3.2x  & 3.7x  & 3.6x        \\ \hline
Parallel Ensemble Distillation & 1.83x & 1.8x  & 1.67x       \\ \hline
\end{tabular}
\end{adjustbox}
\caption{Summary of speedups obtained with our extensions to \emph{FedLess} (\S\ref{sec:extfedless}) for the different datasets.}
\label{table:speedupsummary}
\shrinkspace
\vspace{-6mm}
\end{table}

Tables~\ref{table:fedmdserverlesscost} and~\ref{table:feddfserverlesscost} present the total execution times and costs for the serverless implementations of the two algorithms. For the MNIST and CIFAR datasets with \texttt{FedMD}, we observe that most of the total costs are due to the digest step, which is executed on each client for every round. For the Shakespeare dataset, we observe a significant increase in costs primarily driven by the longer training duration in the revisit step. In contrast to \texttt{FedMD}, we observe comparatively lower costs for \texttt{FedDF}. This is because we only select a fraction of clients, i.e., ten, to participate in each training round using our intelligent selection algorithm (\S\ref{sec:infrasetup},\S\ref{sec:severlessfeddf}). To summarize, \texttt{FedDF} demonstrates cost savings of $34$\% and $82.7$\% compared to \texttt{FedMD} for the MNIST and Shakespeare datasets, respectively, while incurring approximately 16\% higher costs for the CIFAR dataset. 
The higher costs associated with the CIFAR dataset in \texttt{FedDF} can be attributed to the utilization of the entire CIFAR-100 dataset for the ensemble distillation process leading to a higher number of local knowledge distillation steps in every communication round, in contrast to the use of a subset in \texttt{FedMD} (\S\ref{sec:heterodata}).

\subsection{Comparing \texttt{FedDF} and \texttt{FedMD}}
\label{sec:compquantfeddf}

Table~\ref{table:comparision_max_accuracy} shows the maximum Top-1 model accuracy for the two algorithms across all datasets and heterogeneous client model architectures. For the MNIST and CIFAR datasets, we present results with \(\alpha=1\), while for the Shakespeare dataset, we present results for the non-IID data partition. We chose \(\alpha=1\) since it represents a standard non-IID scenario and enables us to compare the robustness of the algorithms toward data heterogeneity. The model architecture level comparison is fair since we use the same architectures for both algorithms. For MNIST, \texttt{FedDF} exhibits higher accuracy levels across all model types compared to \texttt{FedMD}, with an average performance improvement of $5$\% across the six unique model architectures. Similarly, for the CIFAR dataset, \texttt{FedDF} generally outperforms \texttt{FedMD}, except for model 1, where \texttt{FedMD} exhibits better performance. However, on average, across the five unique model architectures, \texttt{FedDF} leads to $3.2$\% better accuracy. Finally, for Shakespeare, \texttt{FedDF} consistently outperforms \texttt{FedMD} by an average of $5$\% across the three unique model architectures.

\vspace{-2mm}
\subsection{Effect of performance optimizations}
\label{sec:perfoptimizationseffect}


Table~\ref{table:speedupsummary} presents summarized results for speedups obtained with our optimizations compared to the original sequential implementation of the two algorithms (\S\ref{sec:extfedless},\S\ref{sec:femd},\S\ref{sec:fedf}). To ensure a fair comparison,  we execute the original algorithms after migrating them to the serverless paradigm. For the initial transfer learning process in \texttt{FedMD} (\S\ref{sec:severlessfedmd}), we obtain an average speedup of $3.5$x across all datasets with our implementation using \texttt{Ray}. On the other hand, for the ensemble distillation process using multiple aggregators in \texttt{FedDF} (\S\ref{sec:severlessfeddf}), we observe an average speedup of $1.76$x across all datasets.



\vspace{-2mm}

\section{Conclusion and Future Work}
\label{sec:concfuture}
In this paper, we took the first step towards training heterogeneous client models in FL via KD using the serverless paradigm. Towards this, we proposed novel optimized serverless workflows for two popular conventional KD techniques, i.e., \texttt{FedMD} and \texttt{FedDF}. To enable adoption, we integrated the two strategies into an open-source serverless FL system called \emph{FedLess} by extending it.  With our experiments, we successfully demonstrated that the serverless implementations of the two strategies converge for heterogeneous model architectures across multiple datasets. In the future, we plan to explore the suitability of more advanced data-free knowledge distillation algorithms that do not require a separate public transfer dataset in serverless environments.

\vspace{-3mm}

\section{Acknowledgement}
\label{sec:ack}
The research leading to these results was funded by the German Federal Ministry of Education and Research (BMBF) in the scope of the Software Campus program under the grant agreement 01IS17049. 

\vspace{-3mm}

\bibliographystyle{ACM-Reference-Format}
\bibliography{serverless} 

\end{document}